\documentclass[pdflatex,sn-mathphys-num]{sn-jnl}

\usepackage{graphicx}%
\usepackage{multirow}%
\usepackage{amsmath,amssymb,amsfonts}%
\usepackage{amsthm}%
\usepackage{mathrsfs}%
\usepackage[title]{appendix}%
\usepackage{xcolor}%
\usepackage{textcomp}%
\usepackage{manyfoot}%
\usepackage{booktabs}%
\usepackage{algorithm}%
\usepackage{algorithmicx}%
\usepackage{algpseudocode}%
\usepackage{listings}%
\usepackage{cleveref}  
\usepackage{xr}        

\theoremstyle{thmstyleone}%
\theoremstyle{thmstyletwo}%

\theoremstyle{thmstylethree}%
\begin{document}

\title[Article Title]{Model-free front-to-end training of a large high performance laser neural network}


\author*[1]{\fnm{Anas} \sur{Skalli}}\email{anas.skalli@femto-st.fr; anasskalli2@gmail.com}

\author[2]{\fnm{Satoshi} \sur{Sunada}}

\author[1]{\fnm{Mirko} \sur{Goldmann}}

\author[3]{\fnm{Marcin} \sur{Gebski}}

\author[4]{\fnm{Stephan} \sur{Reitzenstein}}

\author[4]{\fnm{James A.} \sur{Lott}}

\author[3]{\fnm{Tomasz} \sur{Czyszanowski}}

\author[1]{\fnm{Daniel} \sur{Brunner}}

\affil*[1]{Institut FEMTO-ST,  Université Marie et Louis Pasteur, CNRS UMR, 6174, Besan\c{c}on, France}
\affil[2]{Faculty of Mechanical Engineering, Institute of Science and Engineering, Kanazawa University, Kakuma-machi Kanazawa, Ishikawa 920–1192, Japan}

\affil[3]{Institute of Physics, Lodz University of Technology, ul. Wólczanska 219, 90-924 Lodz, Poland}
\affil[4]{Technical University of Berlin, Hardenbergstra{\ss}e 36, D-10623 Berlin, Germany}


\abstract{Artificial neural networks (ANNs), have become ubiquitous and revolutionized many applications ranging from computer vision to medical diagnoses. However, they offer a fundamentally connectionist and distributed approach to computing, in stark contrast to classical computers that use the von Neumann architecture. This distinction has sparked renewed interest in developing unconventional hardware to support more efficient implementations of ANNs, rather than merely emulating them on traditional systems. Photonics stands out as a particularly promising platform, providing scalability, high speed, energy efficiency, and the ability for parallel information processing. However, fully realized autonomous optical neural networks (ONNs) with in-situ learning capabilities are still rare. In this work, we demonstrate a fully autonomous and parallel ONN using a multimode vertical cavity surface emitting laser (VCSEL) using off-the-shelf components. Our ONN is highly efficient and is scalable both in network size and inference bandwidth towards the GHz range. High performance hardware-compatible optimization algorithms are necessary in order to minimize reliance on external von Neumann computers to fully exploit the potential of ONNs. As such we present and extensively study several algorithms which are broadly compatible with a wide range of systems. We then apply these algorithms to optimize our ONN, and benchmark them using the MNIST dataset. We show that our ONN can achieve high accuracy and convergence efficiency, even under limited hardware resources. Crucially, we compare these different algorithms in terms of scaling and optimization efficiency in term of convergence time which is crucial when working with limited external resources. Our work provides some guidance for the design of future ONNs as well as a simple and flexible way to train them.}




\maketitle
\newpage
\tableofcontents
\newpage
\section{Introduction}

Over the past decade, artificial neural networks (ANNs) have transformed numerous fields, ranging from natural language processing \cite{achiam2023gpt,vaswani2017attention} and image recognition \cite{ard2022five}, to autonomous driving \cite{badue2021self} and advanced game strategies \cite{silver2018general}. Perhaps the best indicator of the success of ANNs and other machine learning algorithms in general, is that they have become so ubiquitous that they are now part of our daily lives, often without us even realizing it.
Their success was the product of a convergence of several key factors: the availability of vast datasets, the rise of affordable and powerful computational resources, and their remarkable ability to tackle complex tasks without requiring explicit programming, making their flexibility unparalleled.\par
ANNs represent a different computing paradigm, grounded in a connectionist and parallel approach to information processing, where a network learns to adapt to perform a task rather than being explicitly programmed to do so. As such, they are fundamentally different from the traditional von Neumann computers used to emulate them.

This mismatch, coupled to fundamental physical properties of electronics, results in significant additional overhead when running ANN algorithms on conventional electronic hardware. In addition, the rapid growth in size and complexity of ANNs has led to a surge in computational demands which has outpaced improvements on the computing front.
This has driven the rapid development and adoption of more parallel von Neumann architectures, such as graphical processing units (GPUs) and tensor processing units (TPUs) \cite{reuther2020survey}. Yet even these are starting to show their limits. Consequently, there has been a surge in interest in researching unconventional hardware platforms to efficiently implement ANNs \cite{hooker2021hardware}. Motivated by these fundamental physical limitations, both academia and industry have been exploring unconventional physical computing substrates that harness nonlinear effects for continuous, analog,  in-memory computing~\cite{jaeger2021towards}.

Optics has re-emerged as a highly promising platform \cite{mcmahon2023physics,abreu2024photonics}, offering scalability \cite{dinc2020optical,rafayelyan2020large,moughames2020three}, high-speed performance \cite{shen2017deep,brunner2013parallel,chen2023deep}, energy efficiency \cite{miller2017attojoule}, and the ability for parallel information processing \cite{brunner2013parallel,lupo2023deep}. Recent advancements in photonic hardware include integrated tensor cores on-chip \cite{feldmann2021parallel} and high-dimensional optical pre-processors \cite{wang2023image,xia2023deep}. Notably, semiconductor lasers are promising candidates for implementing optical neural networks (ONNs) due to their ultra-fast modulation rates and complex dynamics \cite{brunner2013parallel}. Vertical-cavity surface-emitting lasers (VCSELs), in particular, stand out for their efficiency, speed, intrinsic nonlinearity, and well-established CMOS-compatible fabrication process \cite{muller20111550,vatin2018enhanced}.\par

A very popular approach when pursuing unconventional computing is reservoir computing (RC)\cite{jaeger2001echo,maass2002real}, which simplifies recurrent neural networks by transforming input data through a fixed, high-dimensional network of nonlinear nodes—the reservoir—and training only the output weights. This minimal complexity, closely related to extreme learning machines\cite{ortin2015unified}, has enabled implementations on a wide array of physical substrates ranging from electronics~\cite{appeltant2011information} and spintronics~\cite{markovic2019reservoir} to various optical approaches—time-multiplexed~\cite{brunner2013parallel,appeltant2011information}, frequency-multiplexed~\cite{lupo2023deep}, spatially multiplexed~\cite{skalli2022computational,porte2021complete}—and even mechanical systems~\cite{nakajima2013soft}. Recent explorations also include quantum RC to exploit the exponential scaling of Hilbert space for information processing~\cite{markovic2020quantum}.\par

While RC offers flexibility and value through its simplified design, researchers are increasingly exploring fully trainable and more complex architectures to push performance boundaries. In pursuit of harnessing unconventional ANN hardware for specific tasks, significant advances in hardware-compatible training algorithms have emerged~\cite{wright2022deep,momeni2024training,momeni2023backpropagation,nakajima2022physical,xue2024fully}. These developments include model-based techniques such as backpropagation using a digital twin~\cite{wright2022deep} and augmented direct feedback alignment~\cite{nakajima2022physical}, as well as experimental methods for implementing in-situ backpropagation~\cite{xue2024fully, pai2023experimentally}. Simultaneously, model-free or black-box strategies are gaining traction~\cite{skalli2022computational, andreoli2020boolean, momeni2024training, mccaughan2023multiplexed}. However, fully realized in-situ hardware implementations of these methods remain scarce. For unconventional neural networks to be truly competitive, integrating in-situ training techniques is essential to overcome existing bottlenecks and reduce reliance on external high-performance computers—a dependency that often undermines the inherent benefits of these novel computing substrates.\par

In our earlier works \cite{porte2021complete,skalli2022computational,porte2021complete}, we demonstrated RC via spatial multiplexing of modes on a large-area VCSEL (LA-VCSEL), establishing a parallel, autonomous network that minimizes reliance on external computing by employing hardware learning rules with Boolean weights. In this work, we investigate different ways by which we can improve its performance. As such, in section \ref{sec:ceiling} we conduct a ceiling analysis on a software ANN of representative size to understand which parts of the ANN are most crucial to performance. Among these, we find that the presence of both positive and negative weights, a tunable input connectivity as well as sufficient weight resolution are all crucial parameters. This analysis will influence our subsequent ONN designs. The changes to our physical ONN implied by the ceiling analysis require us to rethink the way we train our network. We therefore give a brief overview of the methods used to train physical neural networks in section \ref{sec:strat_survey}. Next in section \ref{sec:strategies}, we present different hardware-compatible, model-free training algorithms in detail. We use them to optimize a toy-problem and to train a software ANN in order to compare them in terms of performance and convergence efficiency. The different evaluated algorithms include: perturbative
methods such as finite difference and simultaneous perturbation stochastic approximation (SPSA), evolutionary algorithms such as CMA-ES and parameter-exploring policy gradients (PEPG), as well as particle swarm optimization (PSO). Among these algorithms, PEPG, CMA-ES and SPSA proved to be the most promising in terms of accuracy and convergence on MNIST, while PSO showed less effective results, struggling to converge efficiently. \par

In section \ref{sec:new_setup} we apply all the conclusions from the ceiling analysis to alter the physical design of our VCSEL-based ONN, going beyond the RC architecture by implementing trainable input weights as well as implementing positive and negative weights with 8-bit resolution using spatial light modulators. We then use the training algorithms studied previously to optimize our ONN . We start with an extensive hyperparameter study of these algorithms in section \ref{sec:hardware_HP_scan} giving guidance and intuition regarding the setting of said hyperparameters. Additionally, we benchmark the different training algorithms using the well-known MNIST hand written digit dataset. Here, rather than simply focusing on pure accuracy, we also study the convergence efficiency of these algorithms in terms of time in section \ref{sec:alg_compare}. Indeed, the time spend to optimize the network is linearly proportional to its energy consumption. This is of particular interest when energy efficiency is crucial, particularly when it comes to online learning under limited hardware resources, which is extremely relevant. Indeed, in physical computing, optimization typically focuses on the efficiency of the hardware, often overlooking the energy impact of the computations required for its training. However, some algorithms, being too resource-intensive, can undermine the expected energy savings. By evaluating various algorithms in terms of both efficiency and computational cost, we introduce a useful perspective on optimizing hardware neuromorphic networks. This approach, which extends beyond the realm of optics, is applicable to a wide range of physical systems, from electronic circuits to quantum devices. \par
We demonstrate a scalable ONN both in size and bandwidth reaching up to 10 000 neurons in parallel, which will be capable of GHz inference due to the fast response time of the laser. We find that PEPG is the most efficient algorithm offering the best performance and convergence efficiency. Interestingly CMA-ES while a widely popular algorithm offers similar performance to PEPG but its heavy quadratic overhead significantly hampers its efficiency in high-dimensional optimization scenarios taking up to 4 times longer to converge in our ONN. SPSA while deceptively simple, offers a compromise between the two algorithms. Finally, we compared the performance of PEPG and SPSA for all classes of the MNIST dataset and compared them to two baselines: a digital linear classifier and a hardware linear system (ONN with LA-VCSEL off). PEPG surpassed the digital linear classifier, while SPSA approached but did not exceed it highlighting the almost unfair advantage of offline weights. Importantly, both algorithms significantly outperformed the hardware linear system, showcasing the high-dimensional transformations enabled by the LA-VCSEL. \par

\section{Ceiling Analysis: Moving Beyond Reservoir Computing and Extreme Learning Machines}
\label{sec:ceiling}
In our previous work \cite{skalli2022computational,skalli2025annealing,porte2021complete} we have shown that a simple LA-VCSEL coupled with information injection and optimized through optically implemented readout weights can be used to solve simple classification tasks. We also studied the influence of physical hyperparameters on performance. As a proof of concept, we chose the classification of binary patterns for their simplicity and tunability. We would now like to move beyond this simple task and show that our system can move towards more complex and "real world" tasks such as the MNIST dataset. To that end, we start by setting the scene by conducting a ceiling analysis on a software NN to know which parts of the NN are most crucial for performance. We then motivate and explain several hardware-compatible training strategies and benchmark them on toy and benchmark problems, paving the way for their future use in our ONN.

\subsection{Motivation and Methodology}

\begin{figure}[h!]
    \begin{center}
    \includegraphics[width=1\linewidth]{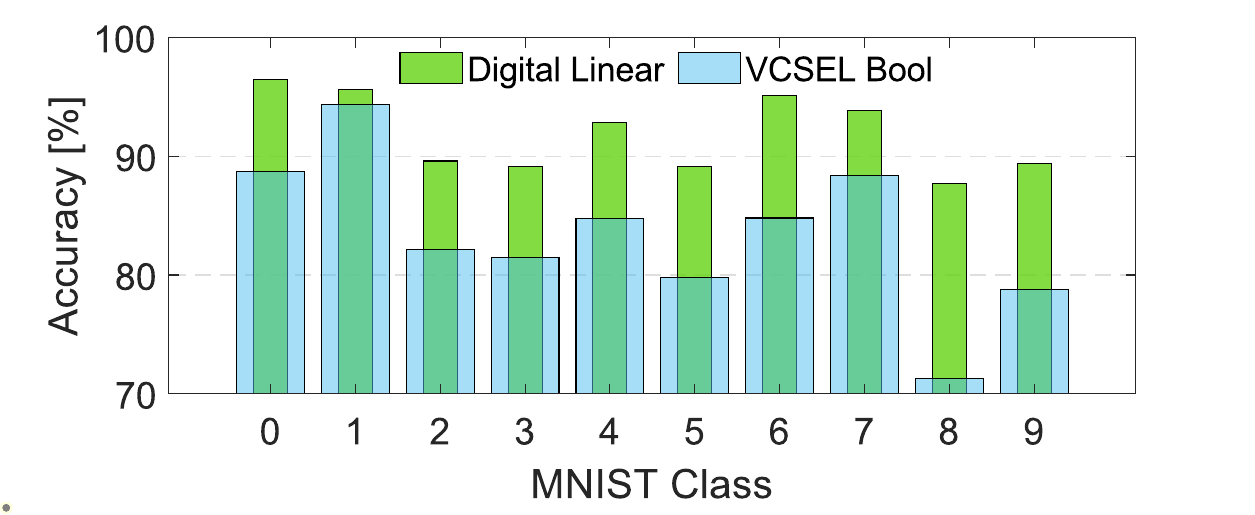}
    \caption{Results of the MNIST classification task using Boolean readout weights of the LA-VCSEL.}
    \label{MNIST_bool_lin}
    \end{center}
    \end{figure}

Let us first benchmark our physical system on the MNIST dataset. This requires no physical change to the experiment, and just requires us to display input sequences comprised of MNIST digits on the input DMD. Using our improved learning strategy, we optimize the Boolean readout on $\text{DMD}_{b}$. Figure \ref{MNIST_bool_lin} shows the results on one-vs-all classification for each MNIST digit and compares it to a simple linear regression model trained and tested on the same experimental data. We use one-vs-all classification as the output of our ONN is a scalar. Each classifier is trained and tested on sequences comprising 1000 images of the target digit and 1000 images of the other digits. Figure \ref{MNIST_bool_lin}, shows that the linear model outperforms the LA-VCSEL for every digit, achieving an accuracy of ${\sim} 92$ \% on average, compared to $84$ \% for the LA-VCSEL.\par

The questions facing us now is how do we improve the performance of our system on this task? Are we limited by hardware factors such as Boolean weights, or our NN architecture, i.e. the fact that we have fixed random input and non-optimized recurrent internal weights? To answer this question, we study the behavior of a software NN with a size comparable to our ONN, i.e. 100 nodes, trained with the back-propagation algorithm. This process is known as conducting a ceiling analysis. By systematically removing or restricting parts of the NN and observing the impact on overall performance, it allows to pinpoint bottlenecks or areas where improvements would have the most significant impact. This method is particularly useful in complex systems like NNs, where it can help us understand how different factors contribute to the system's limitations, guiding future improvements to our ONN.

\subsection{Results of the ceiling analysis}

\begin{figure}[h!]
    \begin{center}
    \includegraphics[width=1\linewidth]{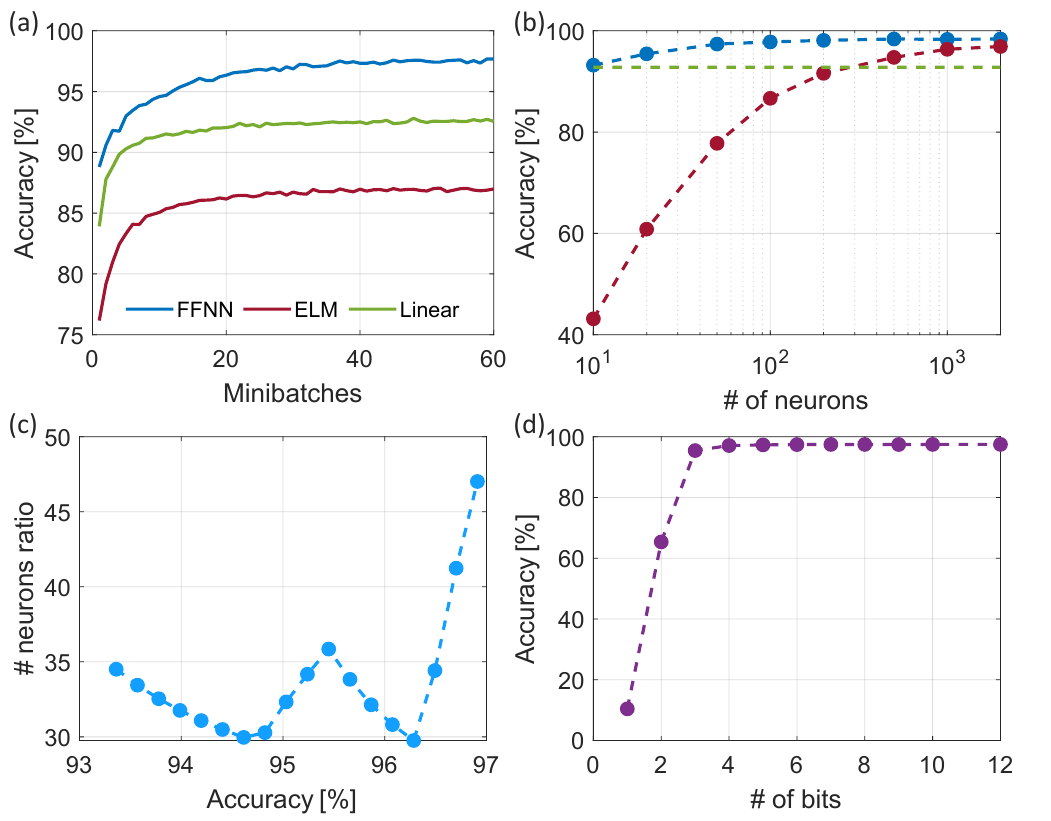}
    \caption{(a) Comparison between a single-layer FFNN in blue and an ELM in red on the MNIST dataset, the baseline performance for a linear classifier is shown in green. (b) Performance for FFNN and ELM architectures as a function of the number of neurons. (c) Ratio between the number of neurons needed for the ELM to reach the same accuracy compared to the FFNN. (d) Performance of the FFNN as a function of the weight resolution.}
    \label{FFNN_ELM_compare}
    \end{center}
    \end{figure}

As shown in Fig. \ref{FFNN_ELM_compare}(a), we first evaluate the performance of a fully connected feed-forward neural network (FFNN) of 100 nodes on the MNIST dataset using Pytorch. This FFNN is fully trained using the backpropagation algorithm with no restriction on the weights. In this reference configuration, the FFNN achieves an accuracy of ${\sim}$97.5\%, as a reminder the accuracy of a linear model is ${\sim}$93\%. If the input weights are fixed, as is the case in our experimental setup, the FFNN becomes an extreme learning machine (ELM) which works similarly to a reservoir in its steady state. The ELM achieves an accuracy of ${\sim}$87\%, showing that the input weights are crucial for performance when working with spatially diverse data. Finally, to simulate a more realistic scenario, we train the FFNN while restricting all weights to be positive only, which proves extremely detrimental to performance, only achieving an accuracy of ${\sim}$60\%. This shows that negative weights are crucial for performance, as they can leverage neurons which are negatively correlated with the target.

To further show the importance of input weights, we study how the performance of, both, FFNN and ELM scale with the number of neurons, as shown by Fig. \ref{FFNN_ELM_compare}(b). By simply training the input weights we are able to leverage more computational capability from the NN. As such, we can decrease the amount of required neurons by a factor of $\approx30\times$ if we implement trainable input weights, see Fig. \ref{FFNN_ELM_compare}(c). Crucially, the ELM only outperforms the linear model when comprising more than 300 neurons. This has considerable implications concerning the use of unconventional physical substrates as means of computing. Indeed, it shows that increasing the number of neurons, which might be difficult and often entails changing substrates, e.g. getting bigger LA-VCSELs, is not the most efficient way to increase performance, while implementing some tunability at the input can lead to better improvements. This raises a potential discussion on the tradeoff between trainable connectivity and nonlinearity in NNs that perhaps deserves a more in-depth study. This tradeoff should in principle be highly task and architecture dependent. Intuitively, we can understand how input weights are crucial to performance when processing spatially diverse data like images. Yet, they may not be as crucial when handling scalar inputs such as some time series. In the latter case, tuning dynamical properties might become more advantageous, and one could for instance study the impact of training internal weights of RNNs. A detailed ceiling analysis for different types of tasks and architectures, focusing on different parts of the NN would be highly interesting, yet is beyond the scope of our study.\par

In the context of a software NN, input weights are potentially trivial to implement and benefit from double precision representation. In contrast, hardware weights would most certainly be limited in resolution, thus we need to study the impact of weight resolution on performance. In the context of optics, 8-bit or even 10-bit resolution is readily achieved by commercially available spatial light modulators (SLM) using liquid crystal on Silicon technology (LCOS). Yet, when it comes to new integrated photonics technologies, such as phase change materials, the resolution is more towards 5-bit \cite{chen2023non}. Figure \ref{FFNN_ELM_compare}(d) shows how performance degrades with lower bit resolution and demonstrates that only marginal improvements in performance are achieved above 4-bit weight resolutions, for a simple single-layer FFNN of 100 neurons applied to the MNIST dataset. In contrast, 2-bit weights cannot solve the MNIST task with the current architecture. It should be noted that quantization of NNs have garnered great interest \cite{guo2018survey,hubara2018quantized} for its memory saving capabilities. Moreover, production models are often trained with full 32-bit resolution gradients and shipped with reduced resolution such as 8-bit to save on memory, especially for NNs running on restricted hardware such as smartphones or microcontrollers \cite{novac2021quantization}. \par

Another crucial aspect of layers in the FFNN is that they implement fully programmable matrix vector operations, which are not trivial to implement in hardware. Indeed, implementing a fully tunable matrix vector multiplication on the input data is non-trivial. In contrast, simply imaging the input data on a SLM yields a Hadamard or element wise product, which can be a good starting point but is vastly inferior in terms of information processing capability.\par

The results discussed thus far are presented in Table \ref{tab:ceiling_analysis}.

\begin{table}[h!]
    \centering
    \begin{tabular}{c|c}
       Architecture  &  Performance\\
       \hline 
         Fully trained, 32-bit & ${\sim}$97.5\% \\
         Fully trained, 4-bit & ${\sim}$97\% \\
         Fully trained positive weights only & 60\% \\
         ELM - training $\text{W}^{\text{out}}$ only & ${\sim}$87\%\\
    \end{tabular}
    \caption{Summary of the main results of the ceiling analysis.}
    \label{tab:ceiling_analysis}
\end{table}

These conclusions may seem trivial in the context of conventional machine learning, yet they have profound consequences when it comes to building a setup that leverages a physical system i.e., the LA-VCSEL as the central piece of an ONN. Firstly, negative weights, while crucial for accuracy, are not trivial to implement optically. Secondly, implementing trainable input weights is a challenge because one cannot rely on error backpropagation to train them as is done in digitally simulated NNs. Therefore, a hardware-friendly optimization algorithm that is sufficiently efficient is required. As a result, we will discuss such optimization algorithms in the next sections.

\section{Survey of hardware-compatible training methods}
\label{sec:strat_survey}

\begin{figure}[h!]
    \begin{center}
    \includegraphics[width=1\linewidth]{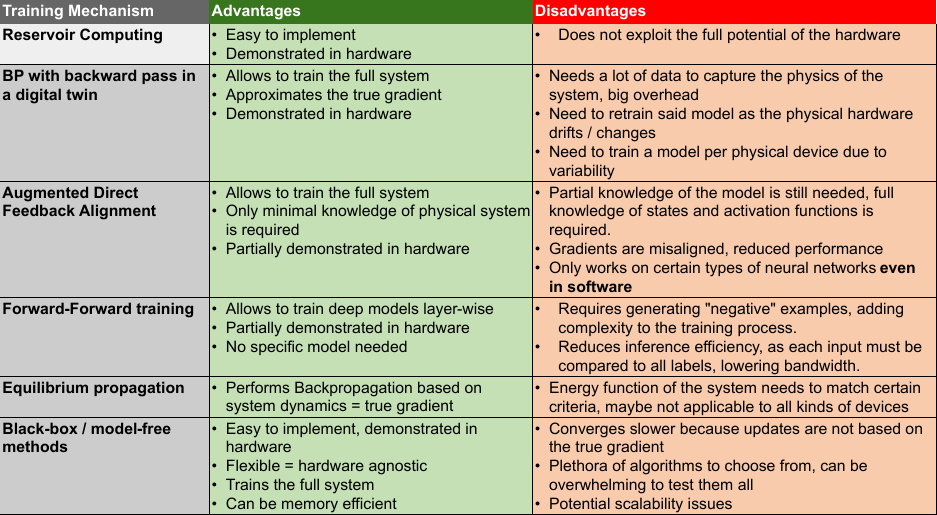}
    \caption{Summary of different hardware compatible strategies with their advantages and disadvantages, compiled from\cite{wright2022deep,nakajima2022physical,hinton2022forward,oguz2023forward,scellier2017equilibrium}.}
    \label{fig:training_methods}
    \end{center}
    \end{figure}

In the context of physical / hardware NNs, online training strategies are usually required to account for physical imperfections such device variability and noise. Figure \ref{fig:training_methods} aims to give a brief summary of the different training strategies that are relevant for hardware NNs, while listing their strengths and weaknesses.

In our case, we want to achieve a fully autonomous NN, and as such we will focus on model-free, or “black-box” training algorithms. Indeed, only model-free methods minimize the reliance on an external computer in contrast to model-based methods e.g. digital twin training \cite{wright2022deep}, which fully relies on a pretrained model to compute the update for hardware weights. 
Moreover, continuous in-the-loop or online learning should compensate for physical system drift, which could hinder the performance of a digital twin over time.
As a result, model-free methods have been singled out as a flexible solution for training physical NNs, yet fully fledged implementations of these algorithms in hardware remain scarce. 
These algorithms can be grouped into two broad categories, on the one hand, perturbative methods with the objective to estimate the gradient by sampling a target function to optimize (i.e. loss function) at different coordinates (i.e. weights). On the other hand, the second class of gradient-free methods consist of population-based sampling, which relies on entirely different concepts and dynamics to optimize functions. These include popular classes of algorithms such as genetic algorithms (GA), evolutionary strategies (ES), swarm optimization and reinforcement learning (RL) algorithms. These are not directly concerned with achieving an approximation of the gradient, but rather iteratively generate better candidate solutions to an optimization problem. They achieve this either according to heuristic criteria in the case of GA, ES, and swarm-type algorithms, or according to an iteratively improved candidate generation policy in the case of RL.
A detailed study comparing the computational cost of training a physical system using model-free versus model-based methods and examining the tradeoff between the high initial cost of pretrained models and the less accurate parameter updates from model-free methods requiring more steps to converge, is still sorely needed at this point but is beyond the scope of this work.

\section{Selected strategies and their implementation}
\label{sec:strategies}
In this section, we present different perturbative and population-based strategies. We will measure the performance of these strategies on optimization problems to showcase their strengths and weaknesses.\par

Our first toy problem is a non-convex function, namely the well-established Rastrigin function, first proposed by Leonid Rastrigin in 1963 \cite{rastrigin1974systems}.
It is defined as follows:

\begin{equation}
    f(x) = An + \sum_{i=1}^{n} (x_i^2 - A\cos(2\pi x_i)).
\end{equation}

\begin{figure}[h!]
    \begin{center}
    \includegraphics[width=1\linewidth]{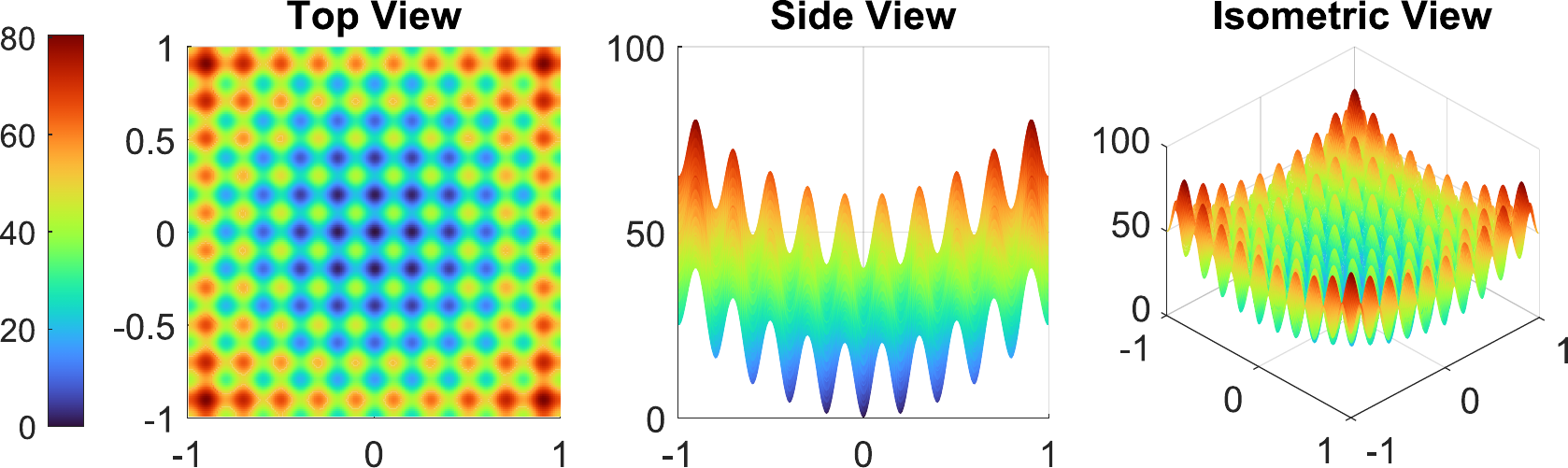}
    \caption{The Rastrigin function in 2D.}
    \label{fig:rastrigin}
    \end{center}
    \end{figure}

This function has a global minimum at $f(x=0)=0$ and is highly non-convex. We will use it to showcase the performance of our training strategies. Figure \ref{fig:rastrigin} shows the Rastrigin function in 2D, with an equal scale for both dimensions, demonstrating the complex nature of the function with many periodically placed local minima. Intuitively, we can see how a gradient-based optimization method could get stuck in the many local minima the Rastrigin function presents. 

Optimization competitions routinely feature the Rastrigin function as a benchmark, benchmark, but they furthermore use a plethora of other functions each one more convoluted than the other to showcase the performance of optimization algorithms. Due to significant progress in the field of optimization, these functions are often combined to create composition functions in order to increase complexity for competitions \cite{liang2013problem}.

Lastly, to provide a more realistic comparison between the different optimization methods, we will use a small NN and train it using the different optimization methods that follow.

\subsection{Gradient measurement perturbative methods}

First, let us discuss the simplest methods. In software NNs, the backpropagation (BP) algorithm is used to determine gradients, subsequently, gradient descent (GD) is then used to update the weights. Although, both, BP and GD are often used interchangeably, they are two distinct algorithms. In hardware, the backpropagation algorithm is not directly implementable for most systems, yet nothing prevents us from implementing a simple GD algorithm, provided we find a way to compute, estimate or measure said gradient, thus we will present two perturbative methods that estimate the gradient.
\subsubsection{Finite difference method}
The simplest estimation of a gradient can be measured by a finite difference scheme. Simply by perturbing a weight by a small amount $\pm \epsilon$ we can measure a difference in error and therefore compute a gradient according to the basic definition of a numerical derivative for a function $L$ around a point $w$ given by

\begin{equation} 
    \frac{\delta L}{\delta w} \simeq \frac{L(w+\epsilon) - L(w-\epsilon)}{2\epsilon}.
\end{equation}

Using this perturbative method on every weight will therefore give us an approximation of the gradient for all parameters, allowing us to perform gradient descent, with the gradient estimate being more accurate for small values of $\epsilon$. Yet, backpropagation makes use of the forward pass of the NN, and can utilize massive hardware vectorization to compute the gradients in a parallel fashion. In contrast, the finite difference method needs to sequentially perturb each weight individually and requires 2 forward passes for each weight. This optimization method while uninspired and has the merit of being local while using very little memory overhead storage as only one weight is perturbed at a time. The pseudocode for an optimization loop using finite difference gradients is given in Algorithm \ref{alg:FD}, describing the optimization of a given function $L$.

    \begin{algorithm}[h]
        \caption{Finite Difference Gradient Descent Optimization}
        \label{alg:FD}
        \begin{algorithmic}[1]
            \State Initialize $\mathbf{w_0}$, set constant learning rate $\alpha$ and a small value for $\epsilon$
            \State $K \gets$ maximum number of iterations
            \State Specify number of parameters to perturb each epoch \(P\)
            \For{$k = 0, 1, \ldots, K-1$}
                \State Initialize gradient vector $\mathbf{g}$ to zeros
                \State Select \(P\) unique parameters randomly from \(\mathbf{w}\) to form subset \(\mathbf{w}_{\text{sub}}\)
                \For{each \(w_j\) in \(\mathbf{w}_{\text{sub}}\)}
                    \State Compute $L^+ \gets L(w_j + \epsilon)$ other parameters stay fixed
                    \State Compute $L^- \gets L(w_j - \epsilon)$  other parameters stay fixed
                    \State \(g_j \gets \frac{L_+ - L_-}{2\epsilon}\) \Comment{Estimate gradient for \(w_j\)}
                \EndFor
                \State Update \(w\) for all \(w_j\) in \(\mathbf{w}_{\text{sub}}\): \(w_j = w_j - \alpha * g_j\)
            \EndFor
            \State \Return Optimized parameters vector $\mathbf{w}_{k_{\text{min}}}$, and optimal Loss value $L(\mathbf{w}_{k_{\text{min}}})$
            \end{algorithmic}
        \end{algorithm}

\subsubsection{Simultaneous Perturbation Stochastic Approximation (SPSA)}

A more sophisticated perturbative method is the Simultaneous Perturbation Stochastic Approximation (SPSA) algorithm. This method was introduced by J. C. Spall in 1987 \cite{spall1987stochastic,spall1992multivariate} and is a stochastic gradient descent based method. It is particularly well suited for hardware-based optimization, as it only requires two function evaluations to estimate the gradient, regardless of the search space dimensionality. It is thus quite a computationally efficient algorithm that has been used to optimize simple NN controllers \cite{choy2004simultaneous,maeda1995learning,martinez2009parameter,wouwer1999use}.

SPSA estimates the gradient vector of a multivariate function by perturbing all parameters $\mathbf{w}$ simultaneously along a randomly chosen direction vector $\Delta$:

\begin{equation}
    \mathbf{g} = \frac{L(\mathbf{w} + \epsilon \mathbf{\Delta}) - L(\mathbf{w} - \epsilon \mathbf{\Delta})}{2\epsilon\operatorname{VAR}(\Delta)} \mathbf{\Delta} \simeq \nabla L(\mathbf{w}),
\end{equation}

where $\mathbf{w}$ is the parameter vector, $\epsilon$ is a small perturbation, and $L$ is the loss function. The direction vector $\Delta$ is chosen to be a random vector with elements $\pm 1$, and $\operatorname{VAR}$ is the variance operator.
We can then use the gradient estimate vector $\mathbf{g}$ to update the parameter vector $\mathbf{W}$ using the standard gradient descent update rule:
\begin{equation}
    \mathbf{w} \gets \mathbf{w} - \alpha \mathbf{g},
\end{equation}
where $\alpha$ is the step size at iteration or learning rate. Naturally, we can use momentum methods or Adam for updates rather than vanilla gradient descent to speed up learning. The pseudocode for an optimization loop using SPSA is given in Algorithm \ref{alg:SPSA}.

\begin{algorithm}[h!]
    \caption{Simultaneous Perturbation Stochastic Approximation (SPSA)}
    \label{alg:SPSA}
    \begin{algorithmic}[1]
    \State Initialize $\mathbf{w_0}$, set constant learning rate $\alpha$ and a small value for $\epsilon$
    \State $K \gets$ maximum number of iterations
    \For{$k = 1, 2, \ldots, K$}
        \State Generate $\Delta_k$, a vector with each element drawn from a Bernoulli $\pm1$ distribution
        \State Compute $L^+ \gets L(\mathbf{w}_k + \epsilon \Delta_k)$
        \State Compute $L^- \gets L(\mathbf{w}_k - \epsilon \Delta_k)$
        \State Estimate gradient: $\mathbf{g}_k \gets \frac{L_+ - L_-}{2\epsilon\operatorname{VAR}(\Delta)} \odot{\Delta}$
        \State Update parameters: $\mathbf{w}_{k+1} \gets \mathbf{w}_k -\alpha *\mathbf{g}_k$
    \EndFor
    \State \Return Optimized parameters vector $\mathbf{w}_{k_{\text{min}}}$, and optimal Loss value $L(\mathbf{w}_{k_{\text{min}}})$
    \end{algorithmic}
    \end{algorithm}

\subsection{Population and sampling based methods}

The second class of optimization methods we explore are so called population-based methods. These methods use information from a population of coordinates to optimize a target function and are as such more global in their nature than the perturbative methods previously introduced. Among these, we can find many algorithms such as the genetic algorithm, different versions of ESs and other nature inspired methods such as particle swarm optimization (PSO), the ant colony algorithm \cite{dorigo2006ant}, the bat algorithm \cite{yang2012bat} and other heuristic methods. ESs are, as the name suggest, a class of "black-box" optimization methods inspired from natural selection with their goal being the optimization of a target function. Before delving into more advanced evolutionary strategies, we lay the general framework for ESs. To develop an intuition for evolutionary strategies, we recommend the illuminating and very well written blog "A visual guide to evolutionary strategies." by D. Ha \cite{ha2017visual}.\par

\subsubsection{Evolutionary strategy basics}

First, let us start conceptually. Following the principle of natural selection, a generation or population of candidates is evaluated with a certain criterion. Following the performance evaluation comes a selection process where only a number of best individuals are selected. These then undergo a recombination process where their best features are combined, with potential random mutation to increase the diversity of candidates. Using this new recombined and mutated population, the next generation of offspring candidates is generated. This new generation embodies the characteristics of their parents and, thanks to their added genetic diversity, some of them will fulfill the selection criterion better than their ancestors. This process of selection, recombination then mutation and finally new generation is repeated, in a "survival of the fittest" loop until a satisfactory condition is met. From this nature-inspired loop let us explain perhaps the simplest yet very illustrative example of an ES. Our goal will be to optimize an objective function $L(\mathbf{w})$. 

First, a random starting position $\mathbf{w_0}$ in the search space is set. Subsequently, a new population of solutions is created around that starting position. Mathematically this is usually done by drawing $p$ sample points $\mathbf{w}_i ,i = \{1, \ldots, p\}$ for a random distribution $\mathbf{w} {\sim} \mathcal{N}\left(\mathbf{w}_{\text{mean}} = \mathbf{w_0}, \sigma^2\right)$, and usually a normal distribution is used. For clarity, we chose $p$ to refer to the population size, usually denoted by $\lambda$, which can be confusing in the context of optics. In practice, the population is simply a list of $p$ points each of which is a candidate solution to our optimization problem. These candidates are subsequently evaluated using the objective function $L(\mathbf{w})$. Following evaluation, a selection process is required, the simplest of which is to choose the best candidate of said population  $\mathbf{w}_{\text{best}} = \arg\min_{\mathbf{w}_i} L(\mathbf{w}_i)$. We then set this best candidate as the new mean for the next generation. In general, for all ESs, increasing $p$ enhances the likelihood of generating better candidate points, albeit at the cost of more sampling. The pseudocode for this simple algorithm is described in Algorithm \ref{alg:SES} \par

\begin{algorithm}
    \caption{Simple ES Optimization}
    \label{alg:SES}
    \begin{algorithmic}[1]
    \State Initialize $\mathbf{w}_{\text{mean}} = \mathbf{w}_0$ randomly, choose $\sigma^2$
    \State Set number of generations $G$ and population size $p$
    \For{$g = 1$ to $G$}
        \For{$i = 1$ to $p$}
            \State Generate $\epsilon_i {\sim} \mathcal{N}(0, \sigma^2)$
            \State Set $\mathbf{w}_i = \mathbf{w}_{\text{mean}} + \epsilon_i$
            \State Evaluate fitness $L(\mathbf{w}_i)$
        \EndFor
        \State Select $\mathbf{w}_{\text{best}} = \arg\min_{\mathbf{w}_i} L(\mathbf{w}_i)$
        \State Update $\mathbf{w}_{\text{mean}} = \mathbf{w}_{\text{best}}$
    \EndFor
    \State \Return $\mathbf{w}_{\text{mean}}$ as the optimized solution
    \end{algorithmic}
    \end{algorithm}

While very simple, Algorithm \ref{alg:SES} lays the framework for more advanced ESs and can be improved with simple tweaks. 
First, instead of selecting only $\mathbf{w}_{\text{best}} = \arg\min_{\mathbf{w}_i} L(\mathbf{w}_i)$ we can selected a subset $\mathbf{W}_{\text{best}}$ of size denoted here by $l$ of best candidates $\mathbf{W}_{\text{best}} = \{\mathbf{w}_{i_1}, \mathbf{w}_{i_2}, \ldots, \mathbf{w}_{i_\mu}\}$ and take their mean for the next iteration $w_{\text{mean}} = \frac{1}{l} \sum_{j=1}^{l} \mathbf{w}_{i_j}$. 
In the literature, $l$ is often denoted by $\mu$, which can lead to confusion when dealing with Gaussian distributions. This variant of the simple ES is usually referred to as the $\mu,\lambda$ (in our case $l,p$) mutation strategy in the literature.\par 
    
We will next present more advanced variants of this basic algorithm that include changes to normal distribution's standard deviation, elongating and changing its orientation to sample more along the direction of the best candidates.

\subsubsection{Covariance Matrix Adaptation Evolution Strategy (CMA-ES)}

Building upon the previously described simple ES, the Covariance Matrix Adaptation Evolution Strategy (CMA-ES or simply CMA) is a strategy that adapts, both, the mean and shape of a multivariate gaussian distribution. As its name suggests, this is done by adapting the covariance matrix of the distribution. It was first introduced in 2001 by N. Hansen and A. Ostermeier \cite{hansen2001completely}. A remarkably well written tutorial has been published by the original authors of CMA in \cite{hansen2016cma}, and the authors also maintain libraries, both, in MATLAB and Python.


The CMA-ES algorithm is analytically quite heavy and as such we will not focus on specific mathematical equations and rather give an intuition for the algorithm. First, the central idea is to adapt a multivariate Gaussian distribution in a space of dimension $D$ for which the general equation is given by:

\begin{equation}
    f(\mathbf{w}) = \frac{1}{\sqrt{(2\pi)^D|\mathbf{C}|}}\exp\left(-\frac{1}{2}(\mathbf{w}-\mathbf{\mu})^T\mathbf{C}^{-1}(\mathbf{w}-\mathbf{\mu})\right),
\end{equation}

\begin{equation}
    \mathcal{N}(\mathbf{\mu}, \sigma^2\mathbf{C}),
\end{equation}
    
where $\mathbf{m}$ is the mean vector, $\mathbf{C}$ is the covariance matrix and $\sigma$ is the step size. We should note that in the context of CMA, $\sigma$ does not refer to the standard deviation of the distribution but is a scalar that is used to determine how far we are from the mean. It is the covariance matrix $\mathbf{C}$ that scales $\sigma$ along certain directions to shape the distribution. Moreover, sigma is adapted independently from $\mathbf{C}$, and it's update depends on the history of the search and, in this sense, it works like a momentum term in optimization algorithms.\par

A simple pseudocode for the CMA-ES algorithm is given in algorithm \ref{alg:CMAES}, below:

\begin{algorithm}
    \caption{Covariance Matrix Adaptation Evolution Strategy (CMA-ES)}
    \label{alg:CMAES}
    \begin{algorithmic}[1]
    \State Initialize population size $p$, number of elites $l$, mean vector $\mathbf{\mu}$, covariance matrix $\mathbf{C}$, step size $\sigma$
    \State Evaluate the initial population based on $\mathbf{\mu}$ and $\mathbf{C}$
    \While{stopping criteria not met}
        \State Generate $p$ new offspring by sampling from $\mathcal{N}(\mathbf{\mu}, \sigma^2\mathbf{C})$
        \State Evaluate fitness of each offspring
        \State Select top $l$ elite offspring based on fitness
        \State Update $\mathbf{\mu}$ towards mean of selected elite
        \State Update $\mathbf{C}$ to reflect distribution of selected elite
        \State Adapt step size $\sigma$ based on success of the search
    \EndWhile
    \State \textbf{return} Best solution found
    \end{algorithmic}
    \end{algorithm}

The CMA-ES algorithm is one of the most advanced ESs and is a widely used algorithm for black-box optimization, in the context of RL or hyperparameter optimization. Yet, it's main drawback is that the computation of the covariance matrix $\mathbf{C}$ is quite heavy and scales with $O(D^2)$. This makes it unsuitable for very high dimensional problems when computational resources are limited.
The original authors proposed a modification simplifying the algorithm and reducing its scaling to $O(D)$, called the Separable CMA-ES (Sep-CMA-ES) \cite{ros2008simple}, which can be seen as only computing the diagonal elements of covariance matrix $\mathbf{C}$. 
    
A simple illustration of the CMA algorithm is provided below in Fig. \ref{fig:CMAES_rastrigin}. For illustration purposes, we start the algorithm at the upper right corner of the Rastrigin function and let it evolve and plot the distribution of the population on top of a contour colormap of the objective function. An additional ellipsoid is superimposed to represent the shape of the distribution. We can see how the ellipsoid first elongates along the direction of best rewards and then gradually contracts as the search converges to the global minimum.

\begin{figure}[h!]
    \begin{center}
    \includegraphics[width=1\linewidth]{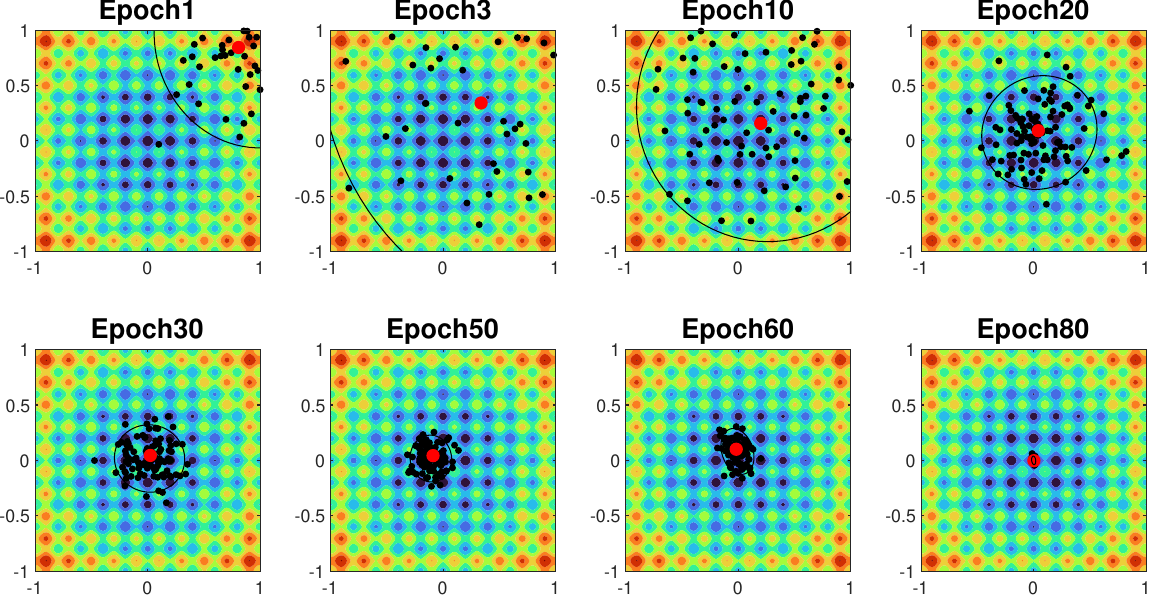}
    \caption{CMA-ES algorithm behavior on the Rastrigin function in $D=10$ dimensions, only the first two are shown. The mean of the distribution is represented by the red dot, the population by the black dots and the ellipsoid helps to visualize the shape of the distribution.}
    \label{fig:CMAES_rastrigin}
    \end{center}
    \end{figure}

\subsubsection{Parameter Exploring Policy Gradients (PEPG)}

The Parameter Exploring Policy Gradients (PEPG) is quite similar to CMA in so far as it is a population-based method that uses information from the population to update the mean as well as standard deviation of said distribution. Yet, instead of modifying the covariance matrix directly, PEPG leverages the policy gradient theorem to estimate the gradient of an objective function with respect to the parameters of its distribution, and then performs gradient descent on the parameters determining said distribution \cite{williams1992simple,sehnke2010parameter}. This concept is used extensively in RL, where optimization problems can be high dimensional and rewards very sparse. In our subsequent explanation, we disregard the technical jargon, as it can be quite confusing and is irrelevant for our purposes.

Let us now explain how PEPG optimization is done in simple terms following the explanation presented in \cite{wierstra2014natural,ha2017visual}. First, we have an objective or fitness function we want to optimize $L(\mathbf{w})$. The central insight is that we will maximize the expected value of $L(\mathbf{w})$ when sampled according to distribution $P_{\theta}(\mathbf{w})$, where $\theta$ represents the parameters of this distribution. For example, it can represent a normal distribution $\mathcal{N}(\mu, \sigma)$, with $\theta = (\mu, \sigma^2)$.
The expected value of $L(\mathbf{w})$ which we want to maximize is given by:

\begin{equation}
    \mathbb{E}[L(\mathbf{w})] = \int L(\mathbf{w}) P_{\theta}(\mathbf{w}) d\mathbf{w}.
\end{equation}

We now rewrite our optimization goal with $\theta$ as our optimization parameter. In simple terms we now have to optimize function $J(\theta) = \mathbb{E}[L(\mathbf{w})]$ with respect to $\theta$. By acting on $\theta$ we influence $P_{\theta}(\mathbf{w})$ and improve the expected value of our fitness function. If the expected value of the fitness function improves, then the best candidate of the population will be even better! Our goal is therefore to perform gradient ascent on $J(\theta)$: 

\begin{equation}
    \theta \gets \theta + \alpha \nabla_{\theta} J(\theta).
\end{equation}

The derivative of $J(\theta)$ is given by: 

\begin{equation}
    \nabla_{\theta} J(\theta) = \nabla_{\theta}\mathbb{E}[L(\mathbf{w})] = \nabla_{\theta}\int  P_{\theta}(\mathbf{w})L(\mathbf{w}) d\mathbf{w}.
\end{equation}

We can permute the gradient $\nabla_\theta$ and the integral $\int d\mathbf{w}$ since $\theta$ and $\mathbf{w}$ are independent and get:

\begin{equation}
    \nabla_{\theta} J(\theta) = \int \nabla_{\theta} P_{\theta}(\mathbf{w})L(\mathbf{w}) d\mathbf{w}.
\end{equation}

We now use a "log derivative trick" given by: 

\begin{equation} 
    \frac{\partial \operatorname{ln}(L(\mathbf{w}))}{\partial \mathbf{w}} = \frac{1}{L(\mathbf{w})}*\frac{\partial L(\mathbf{w})}{\partial \mathbf{w}}. 
\end{equation}

Multiplying by $\frac{P_\theta(\mathbf{w})}{P_\theta(\mathbf{w})} = 1$ inside the integral we get:

\begin{equation}
    \label{eqn:after_log_trick}
    \nabla_{\theta} J(\theta) = \int \frac{P_\theta(\mathbf{w})}{P_\theta(\mathbf{w})}\nabla_{\theta} P_{\theta}(\mathbf{w})L(\mathbf{w}) d\mathbf{w}
    \Leftrightarrow 
    \nabla_{\theta} J(\theta) = \int P_{\theta}(\mathbf{w}) \nabla_{\theta} \operatorname{ln}(P_{\theta}(\mathbf{w}))L(\mathbf{w}) d\mathbf{w}.
\end{equation}

We notice that Eq.\ref{eqn:after_log_trick} can be seen as the expected value of a new variable $ \nabla_{\theta} \operatorname{ln}(P_{\theta}(\mathbf{w}))L(\mathbf{w})$, therefore we have:

\begin{equation}
    \nabla_{\theta} J(\theta) = \mathbb{E}[\nabla_{\theta} \operatorname{ln}(P_{\theta}(\mathbf{w}))L(\mathbf{w})].
\end{equation}

The expectation value $\mathbb{E}[\nabla_{\theta} \operatorname{ln}(P_{\theta}(\mathbf{w}))f(\mathbf{w})]$, can be approximated by sampling a sufficent number $p$ of samples $\mathbf{w}_i$ from $P_{\theta}(\mathbf{w})$: 

\begin{equation}
    \label{eqn:grad_approx}
    \nabla_{\theta} J(\theta) \simeq \frac{1}{p}\sum_{i=1}^{p} \nabla_{\theta} \operatorname{ln}(P_{\theta}(\mathbf{w}_i))L(\mathbf{w}_i).
\end{equation}

Equation \ref{eqn:grad_approx} is known as the policy gradient theorem. Using this theorem, we can now estimate the gradient on $\nabla_\theta J(\theta)$ and optimize our distribution $P_{\theta}(\mathbf{w})$ by acting on its parameters $\theta $ to maximize the expected value of the fitness function $L(\mathbf{w})$. In the case of a normal distribution, $P_\theta(\mathbf{w})=\mathcal{N}_{\theta = (\mu,\sigma)}(\mathbf{w})$. We can therefore update $\mu$ and $\sigma$ using the following update rules:

\begin{equation}
    \mu \gets \mu + \alpha \nabla_{\mu} J(\mu)
    \quad\text{and}\quad 
    \sigma \gets \sigma + \alpha \nabla_{\sigma} J(\sigma).
\end{equation}

For a normal distribution, 

\begin{equation}
    P_\theta(x)=\mathcal{N}_{(\mu,\sigma)}(x) = \frac{1}{\sqrt{2\pi\sigma^2}}\exp\left(-\frac{(x-\mu)^2}{2\sigma^2}\right),
\end{equation}

we obtain, according to\cite{williams1992simple} : 

\begin{equation}
    \nabla_{\mu} \operatorname{ln}(P_{\theta}(\mathbf{w})) = \frac{\mathbf{w}-\mu}{\sigma^2}
    \quad\text{and}\quad 
    \nabla_{\sigma} \operatorname{ln}(P_{\theta}(\mathbf{w})) = \frac{(\mathbf{w}-\mu)^2-\sigma^2}{\sigma^3}.
\end{equation}

The key insight here is that $\nabla_\mathbf{w} L(\mathbf{w})$ is unknown, yet we use the policy gradient theorem to estimate it using $\nabla_{\theta} \operatorname{ln}(P_{\theta}(\mathbf{w}))$. This can be thought of as an optimization by proxy, and works without making any assumptions on $L(\mathbf{w})$ itself. The pseudocode for a simple policy gradient optimization loop is given in Algorithm \ref{alg:PEPG}.


\begin{algorithm}[h!]
    \caption{Simple Policy Gradient optimization}
    \label{alg:PEPG}
    \begin{algorithmic}[1]
    \State Initialize $\theta = (\mu, \sigma)$, set constant learning rate $\alpha$, set population size $p$
    \State $K \gets$ maximum number of iterations
    \For{$k = 1, 2, \ldots, K$} 
        \State Generate $p$ samples $w_i {\sim} \mathcal{N}_{\theta}(\mathbf{w})$
        \State Compute $f(\mathbf{w}_i)$ for each $\mathbf{w}_i$
        \State Estimate gradient: $\nabla_{\mu} J(\mu) \gets \frac{1}{p}\sum_{i=1}^{p} \frac{\mathbf{w}_i-\mu}{\sigma^2}L(\mathbf{w}_i)$
        \State Estimate gradient: $\nabla_{\sigma} J(\sigma) \gets \frac{1}{p}\sum_{i=1}^{p} \frac{(\mathbf{w}_i-\mu)^2-\sigma^2}{\sigma^3}L(\mathbf{w}_i)$
        \State Update parameters: $\mu \gets \mu + \alpha \nabla_{\mu} J(\mu)$
        \State Update parameters: $\sigma \gets \sigma + \alpha \nabla_{\sigma} J(\sigma)$
    \EndFor
    \State \Return Optimized parameters vector $\theta = (\mu, \sigma)$, and optimal Loss value $L(\theta)$
    \end{algorithmic}
    \end{algorithm}

A notable flaw in the simple version of policy gradient optimization presented in Algorithm \ref{alg:PEPG} is that both gradients are inversely proportional to the standard deviation $\sigma$. As the algorithm gradually converges to the global optimum, the standard deviation will shrink which will cause the gradients to explode. This instability similar to an overshooting oscillation problem can be mitigated by using additional momentum terms and applying a time dependent decay on the standard deviation as introduced in \cite{wierstra2014natural}.

\begin{figure}[h!]
    \begin{center}
    \includegraphics[width=1\linewidth]{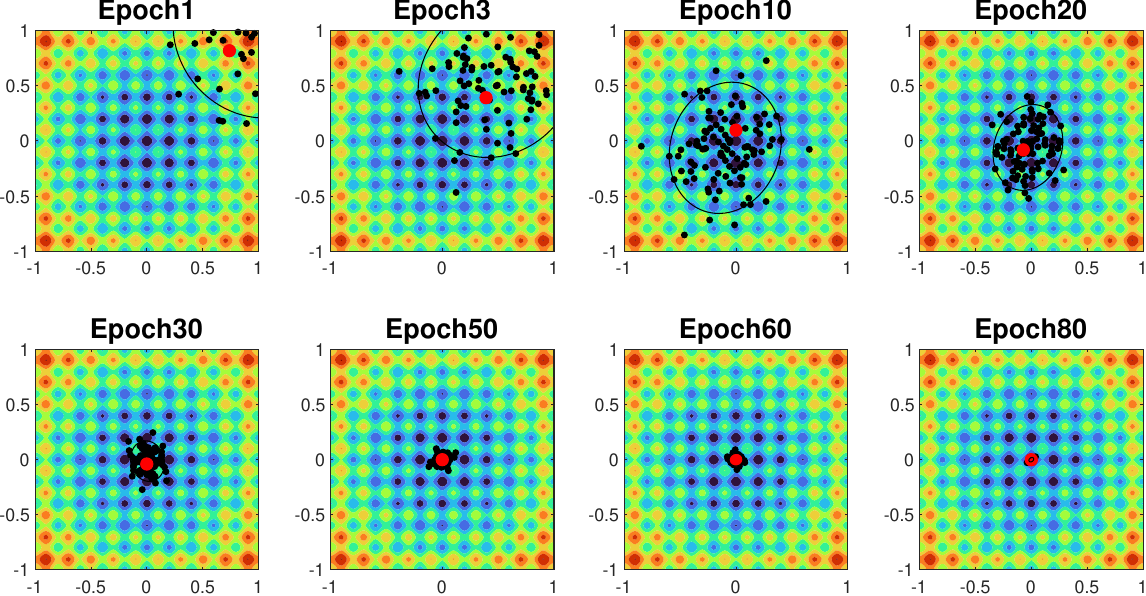}
    \caption{PEPG algorithm behavior on the Rastrigin function in $D=10$ dimensions, only the first two are shown. The mean of the distribution is represented by the red dot, the population by the black dots and the ellipsoid helps to visualize the shape of the distribution.}
    \label{fig:PEPG_rastrigin}
    \end{center}
    \end{figure}

Figure \ref{fig:PEPG_rastrigin} shows the PEPG algorithm in action on the Rastrigin function. We can see how the mean of the distribution is updated towards the global minimum and how the standard deviation is adapted and contracts when the search converges. In practice the distribution remains quite symmetric when using PEPG as opposed to CMA, which is because CMA adapts the off-diagonal terms of the covariance matrix as well. Yet, as PEPG does not compute any covariance matrix, it is more suitable for high dimensional problems as it scales in $O(D)$.

\subsubsection{Particle Swarm Optimization (PSO)}

PSO, unlike the previous methods, is inspired by the behavior and movements of animals such as flocks of birds. As such, it tries to mimic their movements as a way to "patrol" an optimization landscape looking for an optimal solution. Interestingly, the PSO algorithm was introduced by an electrical engineer and social psychologist, Kennedy and Eberhart in 1995 \cite{kennedy1995particle}.

Rather than using a population of candidates and mutation, it uses a swarm of particles, each representing a candidate solution in the search space. Each particle has a velocity, patrols the search space and is attracted to, both, its personal best position and the global best position found by the swarm. These two components influencing the movement of each particle are referred to as the cognitive and social components, the former influencing a more local search, and the latter a more global one. A careful balance between cognitive and social components is needed to efficiently find optimal solutions. Each particle's velocity and position are updated at each iteration of the optimizing loop according to the following equations:

\begin{equation}
    \mathbf{v}^i(t+1) = \omega \mathbf{v}^i(t) + c_1 r_1^i (\mathbf{p}_{\text{best}}^i - \mathbf{w}^i) + c_2 r_2^i(\mathbf{g}_{\text{best}} - \mathbf{w}^i),
\end{equation}
\begin{equation}
    \mathbf{w}^i(t+1) = \mathbf{w}^i(t) + \mathbf{v}^i(t+1),
\end{equation}

where $\mathbf{v}^i$ and $\mathbf{w}^i$ are the velocity and position of particle $i$, $\mathbf{p}_{\text{best}}^i$ its personal best position of particle $i$, $\mathbf{g}_{\text{best}}$ the global best position found by the swarm. All of these being vectors fulfill $\in\mathbb{R}^D$, $D$ being the dimensionality of the search space. $\omega <1$ is the inertia weight and tries to keep the particle on its current trajectory. $c_1$ and $c_2$ are the cognitive and social constants and control how much importance is given to refining the search of a particle $i$ versus acknowledging the best result found by the swarm, and in literature their balancing is often referred to as the tradeoff between 'exploration' and 'exploitation'. $r_1$ and $r_2$ are random numbers drawn from a uniform distribution in $[0,1]^D$. $c_1$, $c_2$, and $\omega$ are hyperparameters that require tuning depending on the optimization problem.

\begin{figure}[h!]
    \begin{center}
    \includegraphics[width=1\linewidth]{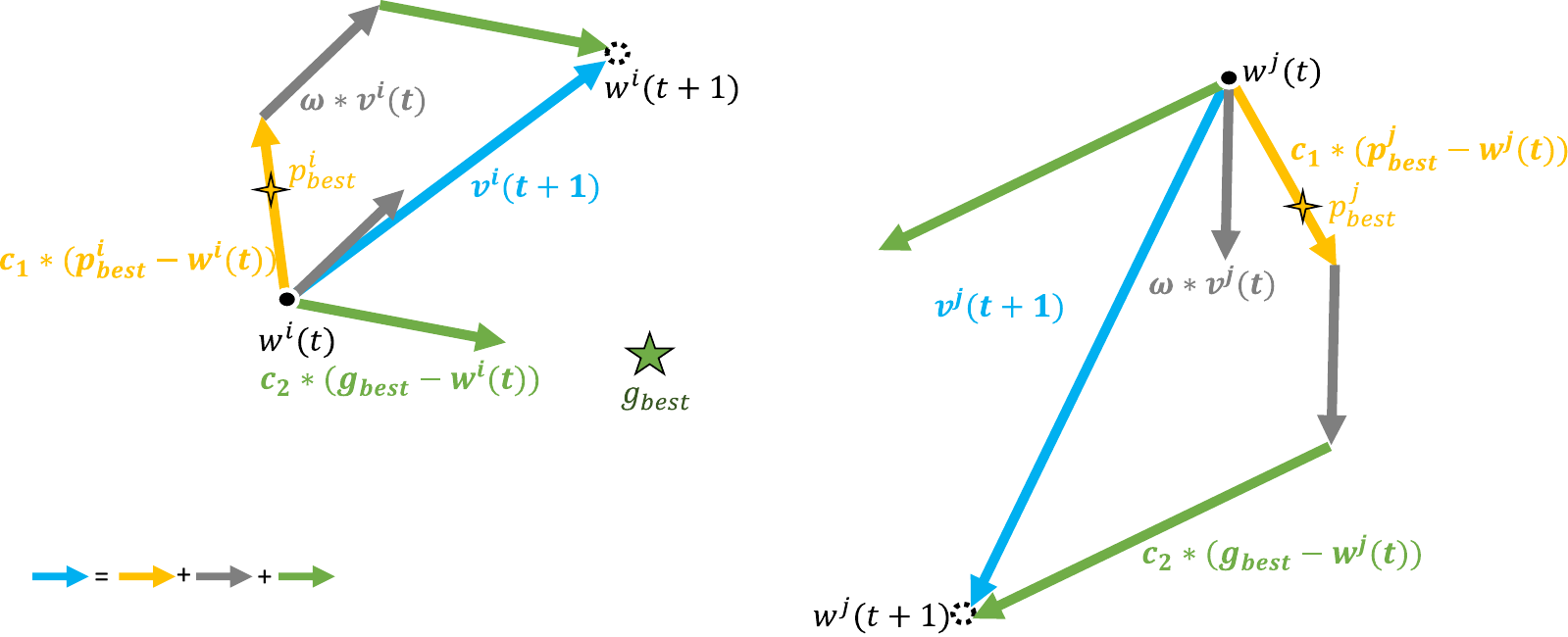}
    \caption{PSO update diagram, showing the update of the velocity of two particles in the swarm $\mathbf{w}^i$ and $\mathbf{w}^j$. The particles are attracted to their personal best $\mathbf{p}_{\text{best}}^i$ and the global best $\mathbf{g}_{\text{best}}$. The inertia weight $\omega$ tries to keep the particle on its current trajectory.}
    \label{fig:PSO_diagram}
    \end{center}
    \end{figure} 

Figure \ref{fig:PSO_diagram} describes the update of the particles' velocity and position. The process can be illustrated by thinking of $(\mathbf{p}_{\text{best}}^i - \mathbf{w}^i) $, $(\mathbf{g}_{\text{best}} - \mathbf{w}^i)$ and $\mathbf{v}^i(t)$ as three vectors scaled by their respective constants $c_1$, $c_2$, and $\omega$ yielding vector $\mathbf{v}^i(t+1)$. The particle then moves in the direction of the sum of these 3 scaled vectors $\mathbf{v}^i(t+1)$. The pseudocode for PSO is given in Algorithm \ref{alg:PSO}.

\begin{figure}[h!]
    \begin{center}
    \includegraphics[width=1\linewidth]{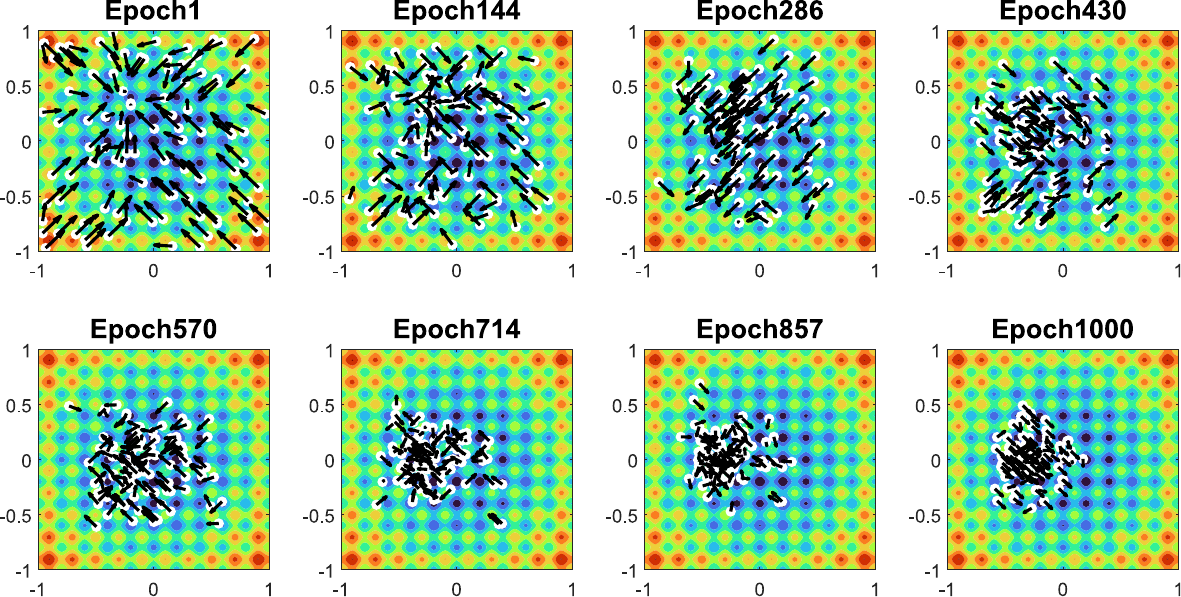}
    \caption{PSO algorithm behavior on the Rastrigin function in $D=10$ dimensions, only the first two are shown. The particles are attracted to their personal best $\mathbf{p}_{\text{best}}^i$ and the global best $\mathbf{g}_{\text{best}}$. The inertia weight $\omega$ tries to keep the particle on its current trajectory.}
    \label{fig:PSO}
    \end{center}
\end{figure}

\begin{algorithm}[h!]
    \caption{Particle Swarm Optimization (PSO)}
    \label{alg:PSO}
    \begin{algorithmic}[1]
    \State Set swarm size $p$, maximum number of iterations $K$.
    \State Initialize each particle $i$ in swarm with random positions $\mathbf{w}^i$ and velocities $v^i$.
    \State Initialize hyperparameters $c_1$, $c_2$, and $\omega$.
    \State Evaluate the fitness of each particle at $\mathbf{w}^i$.
    \State Set personal best $\mathbf{p}_{\text{best}}^i = \mathbf{w}^i$ and global best $\mathbf{g}_{\text{best}}$ to the best $\mathbf{w}^i$ among all particles.
    \While{termination criteria not met}
        \For{each particle $i$ in the swarm}
            \State Update velocity:
            \State $v^i = \omega v^i + c_1 r_1 (\mathbf{p}_{\text{best}}^i - \mathbf{w}^i) + c_2 r_2 (\mathbf{g}_{\text{best}} - \mathbf{w}^i)$
            \State Update position:
            \State $\mathbf{w}^i = \mathbf{w}^i + v^i$
            \State Evaluate the new fitness of particle $i$ at $\mathbf{w}^i$.
            \If{new fitness is better than $\mathbf{p}_{\text{best}}^i$}
                \State Update $\mathbf{p}_{\text{best}}^i$ to the new position $\mathbf{w}^i$.
            \EndIf
            \If{new fitness is better than the fitness of $\mathbf{g}_{\text{best}}$}
                \State Update $\mathbf{g}_{\text{best}}$ to $\mathbf{w}^i$.
            \EndIf
        \EndFor
    \EndWhile
    \State \Return $\mathbf{g}_{\text{best}}$.
    \end{algorithmic}
    \end{algorithm}

    PSO has been used for various optimization problems such as training NNs\cite{khan2012comparison,rauf2018training}, reservoir computers \cite{wang2015optimizing} as well architecture search \cite{wang2020particle} where the architecture and hyperparameters of NNs are optimized. PSO main advantage is its propensity towards global search as the swarm is uniformly initialized in the search space. Yet, it can often result in a premature convergence to a local minimum. Figure \ref{fig:PSO} shows the convergence behavior of the PSO algorithm on the Rastrigin function. At first, individual particles are uniformly distributed and each of them has a given velocity and direction, the swarm-like behavior is quite apparent and over time particles tend to converge and group together to offer a more focused search. Yet, as shown in Fig. \ref{fig:PSO}, the algorithm is unable to find to global minimum and prematurely converges to a local one.

\clearpage

\subsection{Application to Toy problems}

\subsubsection{Objective Functions}

\begin{figure}[h!]
    \begin{center}
    \includegraphics[width=1\linewidth]{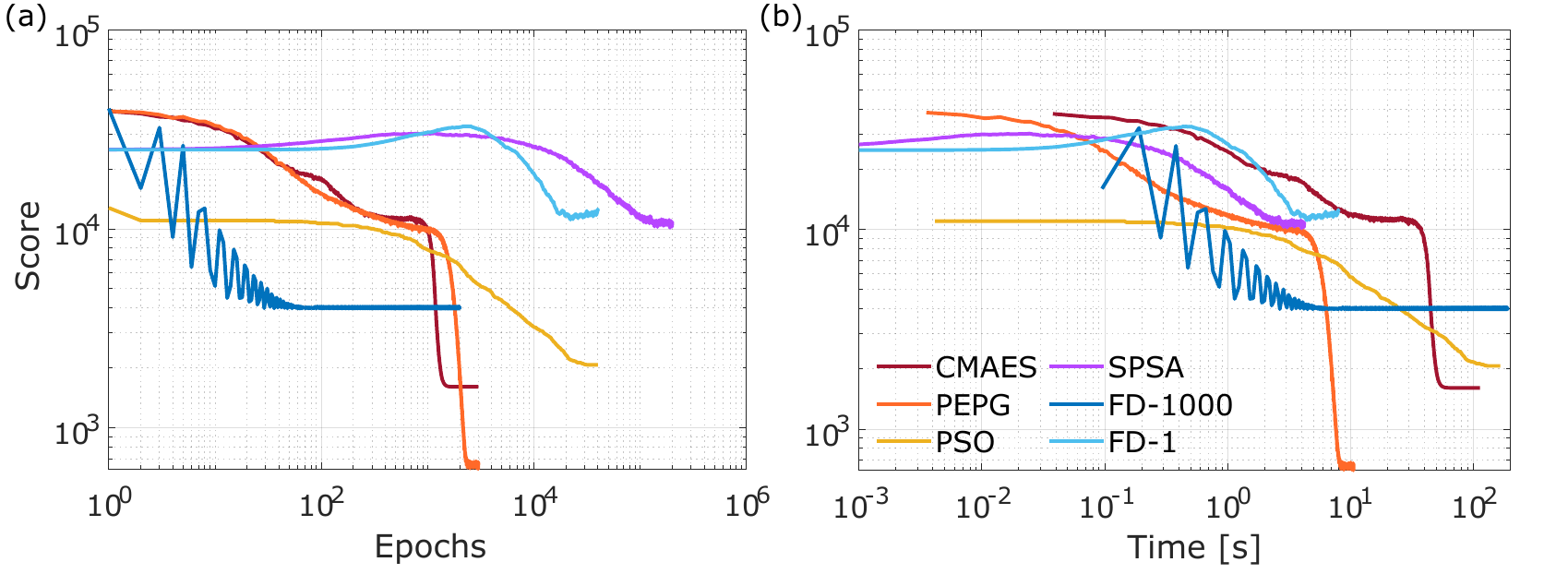}
    \caption{Performance on the Rastrigin function for each optimization algorithm as a function of
    epochs (a) and of time in seconds (b).}
    \label{fig:score_rastrigin}
    \end{center}
    \end{figure}

Let us start by benchmarking the optimization algorithms on the Rastrigin function in $D=1000$ dimensions, which offers a significant challenge, but is still quite low dimensional compared to problems like NN optimization. For population-based methods, we use a population size of 100. Figure \ref{fig:score_rastrigin}(a) shows the so called score i.e. minimal value of the Rastrigin function achieved by each algorithm at every epoch i.e. iteration of the algorithm, while panel (b) shows the same information but now plotted against true time in seconds. This therefore includes the computational overhead of each algorithm.\par

The general trend is that population-based methods achieve lower values of the Rastrigin function than perturbative methods. Perturbative gradient estimates methods like FD and SPSA break down for this optimization problem. Indeed, the Rastrigin function is highly symmetric, which makes it hard for gradient based methods to escape local minima even when using momentum-based optimizers like ADAM. When increasing the number of directions along which the gradient is estimated to 1000 the FD method performs significantly better thanks to a more accurate gradient estimate, yet it also converges prematurely to a local optimum. It should be noted that probing all available dimension of an optimization problem independently in a sequential manner would take too long and is not feasible in practice.\par


The CMA-ES and PEPG algorithms are the best performing, with PEPG performing slightly better and converging significantly faster by requiring only $10s$, compared to CMA-ES that requires up to $60s$ even in this relatively low dimensional case. This sharp increase in convergence time for the CMA-ES algorithm stems from the computational overhead presented by the computation of the covariance matrix, and due to the scaling, this difference is exacerbated as the dimensionality of the optimization problem grows. In contrast, the particle swarm algorithm shows poorer performance as it struggles to navigate the landscape in a meaningful way. Even if PSO has the same overhead as PEPG, it ends up converging slower than CMA-ES due to its poor ability to use information from the landscape. Moreover, PSO is the only algorithm that required us to do extensive hyperparameter tuning. Noteworthy is also that it is extremely challenging to develop an intuition for PSO hyperparameters, making their tuning process highly cumbersome.\par

\begin{figure}[h!]
    \begin{center}
    \includegraphics[scale=0.5]{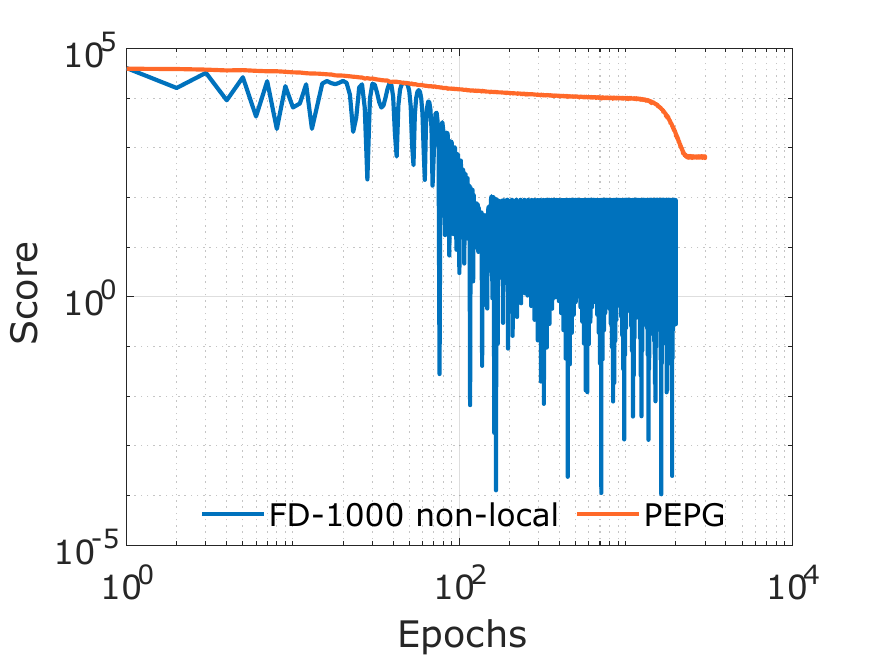}
    \caption{Performance on the Rastrigin function for for PEPG and FD using $\epsilon = 0.1$.}
    \label{fig:cheating_rastrigin}
    \end{center}
    \end{figure}

Although the Rastrigin function is a good toy problem for benchmarking, studying and understanding optimization algorithms, it does not really fit our purposes since it is highly symmetric and is not representative of a real error landscape as would be found in a NN. The highly symmetric nature of the version of the Rastrigin function makes it easy to 'cheat'. Indeed, increasing $\epsilon$ to $\epsilon = 0.1$ in the SPSA or FD algorithm allows it to 'jump' over local minima, leveraging more global information, and yielding significantly better performance than even PEPG as shown in Fig. \ref{fig:cheating_rastrigin}. It should be stressed, however, that this is only possible because the structure of the Rastrigin function is pseudo periodic, moreover the higher $\epsilon$ leads to significant instability. Generally, such a high $\epsilon$ value no longer yields an accurate estimate of the gradient.\par

\subsubsection{Neural network optimization}

\begin{figure}[h!]
    \begin{center}
    \includegraphics[scale=0.5]{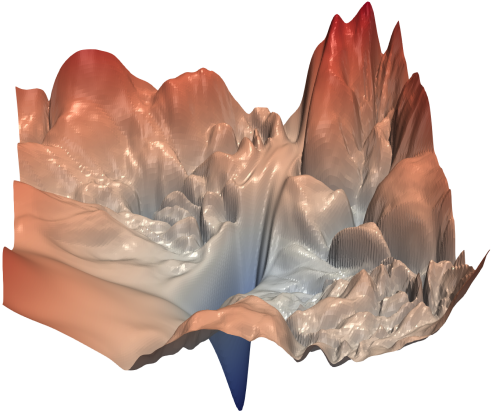}
    \caption{Illustration of the error landscape of a convolutional NN, reproduced from \cite{li2018visualizing}.}
    \label{fig:NN_landscape}
    \end{center}
    \end{figure}

As stated in the previous section, rather than arbitrary optimization functions, a more useful test would be to benchmark all optimization algorithms in the context of NN training. A detailed breakdown of the shape of NN error landscapes is provided in \cite{li2018visualizing}. Figure \ref{fig:NN_landscape} illustrates the rough, irregular and steep landscape of a convolutional NN. Therefore, it becomes quite apparent that an artificial function, such as given by Rastrigin, will only provide us with limited information. \par

As such, we chose 2 classification tasks for a NN to solve, namely the IRIS \cite{fisher1936use} and   MNIST \cite{lecun1989backpropagation} datasets. We chose these two datasets since MNIST provides high dimensional input images with $784$ input features, while the IRIS dataset gives us the limit scenario of a very low dimensional dataset with only $4$ input features. Table \ref{tab:datasets} bellow offers a quick overview of each dataset:

\begin{table}[h!]
    \centering
    \caption{Summary of MNIST and Iris Datasets.}
    \begin{tabular}{|l|c|c|}
    \hline
    \textbf{Dataset Feature} & \textbf{MNIST} & \textbf{Iris} \\
    \hline
    Input Features           & 28x28 = 784 pixel images & 4 flower measurements \\
    \hline
    Output Features          & Digit labels (0-9) & Iris species (setosa, versicolor, virginica) \\
    \hline
    Number of Classes        & 10 & 3 \\
    \hline
    Training Samples         & 60,000 & 105 \\
    \hline
    Testing Samples          & 10,000 & 45 \\
    \hline
    \end{tabular}
    \label{tab:datasets}
    \end{table}

Like most image-based datasets, the MNIST dataset is quite high dimensional with a total of $784$ input features. If we were to consider a FFNN of $100$ nodes, then the number of trainable parameters would total $784*100 + 100 + 100*10 +10 =  79 510$, which include the input weights, node biases, the output weights, and output biases. Such a high number of parameters would make our study highly time-consuming, and would render the CMA-ES algorithm unusable. Therefore, to significantly reduce the computational cost of our analysis and particularly that of the CMA-ES algorithm, we will use a ConvNet architecture of ${\sim}12 000$ parameters on the MNIST dataset. The specific architecture used is presented in table \ref{tab:MNIST_architecture} and is similar to the one used in \cite{ha2017visual}. On the contrary, the IRIS dataset is quite low dimensional with only $4$ input features, and we will use a simple FFNN with $1$ hidden layers of $10$ nodes. The architecture is presented in table \ref{tab:IRIS_architecture}.

\begin{table}[h!]
    \centering
    \small 
    \begin{tabular}{|c|c|c|c|p{2cm}|}
    \hline
    \textbf{Layer} & \textbf{Type} & \textbf{Output Shape} & \textbf{Number of Parameters} & \textbf{Description} \\
    \hline
    Input & Input Layer & (28, 28, 1) & 0 & Takes an input of size 28x28x1 (grayscale image) \\
    \hline
    Conv1 & Conv2D & (28, 28, 8) & 208 & A convolutional layer with 8 filters of size 5x5, stride 1, and padding 2 \\
    \hline
    MaxPool1 & MaxPooling2D & (14, 14, 8) & 0 & A max pooling layer with a 2x2 window and stride 2 \\
    \hline
    Conv2 & Conv2D & (14, 14, 16) & 3216 & A convolutional layer with 16 filters of size 5x5, stride 1, and padding 2 \\
    \hline
    MaxPool2 & MaxPooling2D & (7, 7, 16) & 0 & A max pooling layer with a 2x2 window and stride 2 \\
    \hline
    Flatten & Flatten & (784) & 0 & Flattens the output to a vector of size 784 \\
    \hline
    FC & Fully Connected & (10) & 7850 & A fully connected layer with 10 output neurons \\
    \hline
    Output & Output Layer & (10) & 0 & Produces the final output of size 10 \\
    \hline
    \end{tabular}
    \caption{Summary of the convolutional network architecture used on the MNIST dataset.}
    \label{tab:MNIST_architecture}
    \end{table}

    \begin{table}[h!]
        \centering
        \small
        \begin{tabular}{|c|c|c|c|p{2cm}|}
        \hline
        \textbf{Layer} & \textbf{Type} & \textbf{Output Shape} & \textbf{Number of Parameters} & \textbf{Description} \\
        \hline
        Input & Input Layer & (4) & 0 & Takes an input of size 4 (features of the Iris dataset) \\
        \hline
        Hidden1 & Linear & (10) & 50 & Fully connected layer with 10 output neurons \\
        \hline
        ReLU & ReLU Activation & (10) & 0 & Applies the ReLU activation function \\
        \hline
        Output & Linear & (3) & 33 & Fully connected layer with 3 output neurons (corresponding to the 3 classes of the Iris dataset) \\
        \hline
        \end{tabular}
        \caption{Summary on the FFNN architecture used on the IRIS dataset.}
        \label{tab:IRIS_architecture}
        \end{table}

\begin{figure}[h!]
    \begin{center}
    \includegraphics[width=1\linewidth]{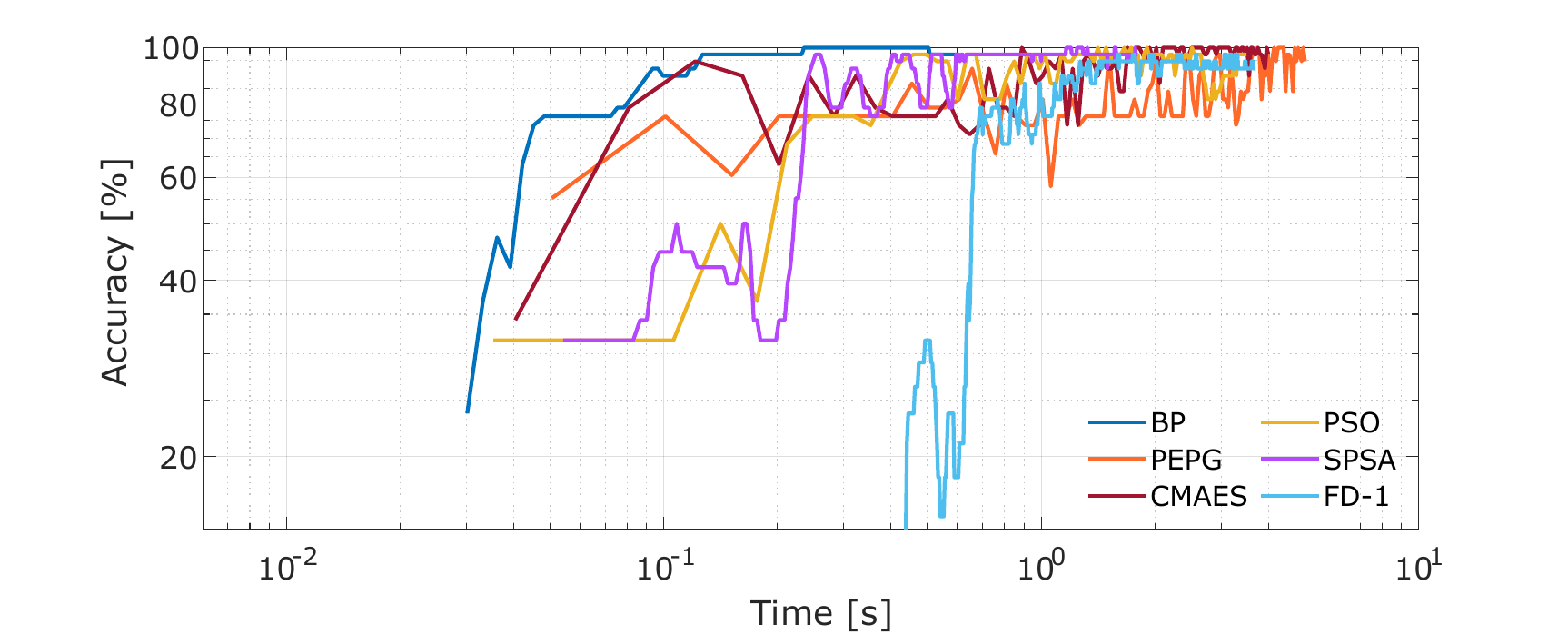}
    \caption{Convergence of the different optimization algorithms on the IRIS dataset.}
    \label{fig:iris_online_simulation}
    \end{center}
\end{figure}

We can then compare all training algorithms in terms of convergence efficiency measured in seconds for the two datasets and compare them with BP for reference. Performance on the IRIS dataset, shown in Fig. \ref{fig:iris_online_simulation} is less interesting and was performed as a sanity check, to see how all algorithms would perform on a low dimensional dataset. As expected, all algorithms can reach $100\%$ classification accuracy in barely more than $1$s. The task is too low dimensional to explicitly reveal any specific shortcomings of these algorithms. \par

\begin{figure}[h!]
    \begin{center}
    \includegraphics[width=1\linewidth]{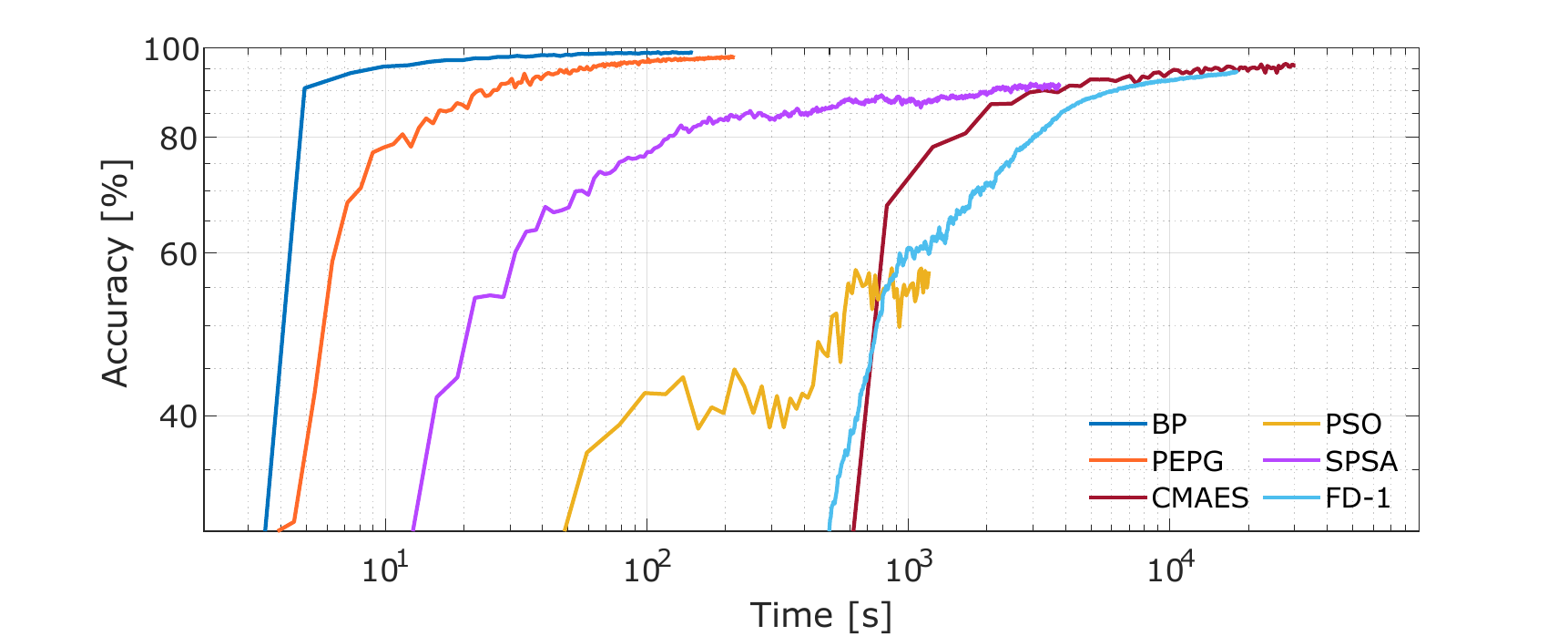}
    \caption{Convergence of the different optimization algorithms on the MNIST dataset.}
    \label{fig:MNIST_online_simulation}
    \end{center}
\end{figure}

Figure \ref{fig:MNIST_online_simulation} showcases performance on the MNIST dataset. Here, BP achieves ${\sim} 99\%$ classification accuracy, which constitutes the ceiling against which we can evaluate other algorithms. BP is followed by PEPG at  ${\sim} 98\%$ and CMA-ES at ${\sim} 96\%$. Crucially, because of its heavy quadratic overhead, each iteration of the CMA-ES algorithm takes a significant amount of time, there is a factor ${\sim} 30$ between updates computed with PEPG and CMA-ES. In general, CMA-ES is seen as more robust and should achieve higher performance than PEPG, yet because of its expensive overhead we are not able to tune its hyperparameters in the present task. In practice, this makes PEPG more robust for high dimensional optimization problems under limited resources. The PSO algorithm performs poorly in general and is not able to converge to a satisfactory solution, it seems that it is not well suited to high dimensional optimization problems. All population-based methods were evaluated with a population size of $200$.\par 

Regarding perturbative methods, it should be stressed that in the context of digital NNs, there is no difference between the true gradient given by BP and a finite difference method that would probe all parameters because $\epsilon$ can be arbitrarily small. Indeed, as a method to debug digital NNs, FD can be used to perform so called "gradient checking" to make sure that computed NN gradients are correct. However, gradient checking has largely become obsolete due to the widespread adoption of automatic differentiation techniques, which no longer require users to explicitly program gradient computations when using conventional NN libraries such as Pytorch or TensorFlow. \par
Surprisingly, SPSA converges significantly faster than FD-1 while having the same overhead, since both probe a single direction. The main difference here is that FD-1 probes a single parameter while SPSA probes a direction that is a linear combination of all parameters, enabling it to use more information and converge quicker. Indeed, since MNIST data is quite sparse, i.e. a significant portion of the image is just a dark background, FD-1 can spend quite some time optimizing irrelevant parameters, unlike SPSA which perturbs all parameters at the same time.

SPSA reaches ${\sim} 92\%$ accuracy on the MNIST dataset, which is quite impressive given its simplicity, while FD-1 also reaches ${\sim} 92\%$ but takes ${\sim} 3$ times longer, and then outperforms SPSA due to its highly accurate estimate of the true gradient. Regarding perturbative methods, it should be noted that $\epsilon$ values such as $\epsilon = 10^{-6}$ are unrealistic in hardware and would require more than $16$ bits of resolution. A more detailed study of this aspect is presented in section \ref{sec:hardware_HP_scan}.\par 

\begin{figure}[h!]
    \begin{center}
    \includegraphics[width=1\linewidth]{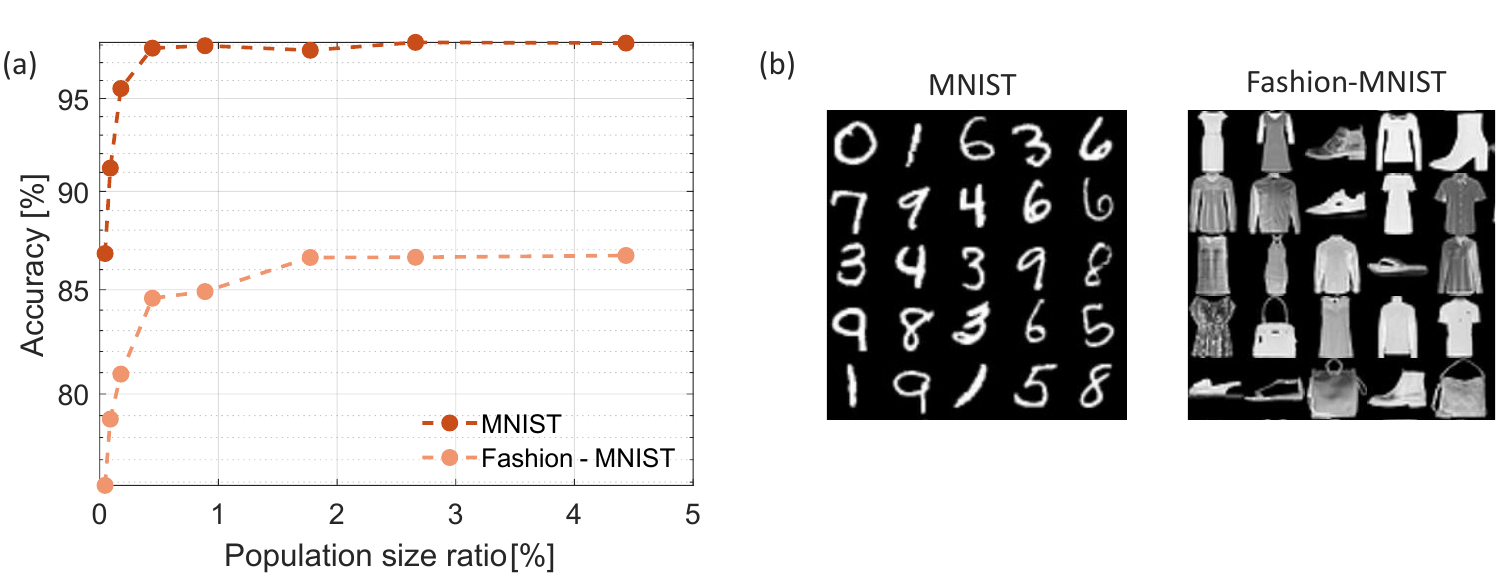}
    \caption{Accuracy of the PEPG algorithm on the MNIST dataset as a function of the population size ratio to the total number of parameters in the network.}
    \label{fig:PEPG_mnist_pop_size}
    \end{center}
\end{figure}

Finally, we can study how performance scales with population size for the best population-based algorithm, PEPG. Figure \ref{fig:PEPG_mnist_pop_size}(a) shows a study of accuracy as a function of the ratio between the population size and the total number of parameters to optimize in the convolutional network for the MNIST dataset. Interestingly, for this task performance saturates at ${\sim} 1\%$ corresponding to a population size of $113$. It should be noted that this ratio is not absolute in any sense and should, in principle, be highly task dependent.\par
Intuitively, we can understand how only a small population size is needed by looking at the MNIST dataset itself, shown in Fig. \ref{fig:PEPG_mnist_pop_size}(b). The images are quite sparse and mostly consist of a dark background, therefore only a small number of pixels corresponding to the digits themselves are actually relevant for classification. One could imagine how with another dataset, such as fashion-MNIST, where the images are fuller and more complex, a larger population size would be needed to optimize the network. Indeed, when using PEPG on the fashion-MNIST dataset, we found that performance saturates for a population size ratio of ${\sim} 2\%$ corresponding $226$ samples, confirming our hypothesis.

To summarize, in these previous sections, we laid the groundwork for future improvements to our ONN experimental setup. These improvements stem from insights given by a ceiling analysis performed on a software NN. This celling analysis led us to three central conclusions. Firstly, having both positive and negative weights is crucial in a NN. Secondly, these weights should have sufficient resolution to account for the complexity of the task. And finally, having trainable input weights as opposed to a fixed random mapping is crucial for performance. These conclusions, while seemingly trivial, have a profound impact on the design of our ONN, particularly when it comes to weight optimization. We therefore presented several training strategies of varying complexity and applied them to, both, toy problems and in the context of NN optimization on the MNIST and IRIS dataset. We were able to verify that several of these algorithms, particularly PEPG, SPSA and CMA-ES show promising performance and varying degrees of overhead cost. In the next sections, we will use these insights to improve our ONN experimental setup and use the hardware compatible algorithms presented here to train a fully tunable ONN on the MNIST dataset.

\newpage
\section{Towards a Fully Tunable Optical Neural Network}
\label{sec:new_setup}
\subsection{Large area VCSELs}

For our proof-of-concept hardware NN implementation, we built an optical reservoir or extreme learning machine for image processing, with a VCSEL playing the starring role. This, in turn, begs the following question, "What type of VCSEL should we use?" Indeed, VCSELs come in various shapes and sizes, yet, as stated previously, we want our NN to process information in a parallel fashion. To achieve that we will use a spatially multimode VCSEL. In this spatially-multiplexed approach, transverse modes of the VCSEL will be related to neurons of the network. In other words, the more modes our VCSEL can support, the larger its potential information processing capabilities. 

As a first approximation the mode profile, i.e. the spatial distribution of light that a semiconductor laser exhibits, results from interplay between two physical effects. First, the geometrical properties of the cavity provide confinement conditions that support certain optical modes. These are solutions to Maxwell's equations inside the medium, and are referred to as "cold cavity" modes, as they do not take into consideration any carrier effects via current injection, i.e. conditions as encountered as the device is pumped. When current is injected into the cavity, charge carriers are created and, via stimulated emission, they provide gain to certain modes of the cold cavity which in turn become lasing modes. This means that in practice, when the device is switched on and imaged on a camera, the VCSEL's emission mode profile is determined by the optical modes that overlap the most adequately with the distribution of carriers in the device. In simple terms, there is a number of carriers available to provide gain and different optical modes "compete" for these carriers. Finally, other physical properties such as temperature and carrier distribution can also influence the refractive index distribution which in turn influences the mode profile.

The multimodal nature of the LA-VCSEL is a central point in our NN design. As a thought experiment, let us consider the case of a single mode VCSEL. For such a device, spatially diverse input data is mapped onto the only mode that the VCSEL supports, making it therefore impossible to spatially distinguish different inputs. Our VCSEL should therefore be sufficiently multimode in order to spatially separate input data. Finally, the multimodal property of the laser is a product of the cavity geometry, namely its dimensions and shape. The following discussion explores various LA-VCSEL designs, highlighting those best suited for our application.

\subsubsection{Circular cavity VCSELs}

Circular cavity VCSELs represent the overwhelming majority of, if not all devices in commercial use today. Two other types of cavities, namely chaotic cavities and photonic crystal cavities, are generally confined to academic research \cite{skalli2022photonic}. Circular cavity VCSELs are robust and well-studied devices, yet because our goal is to build an ONN based on the spatial multiplexing of modes, we need to choose devices which inherently maximize the number of supported spatial modes. In practice, this means choosing bigger devices. For technological applications, circular VCSELs are generally designed to operate in a in a single mode (or near single mode) regime which means the commercial circular oxide aperture VCSELs are quite small, usually under ${\sim} 5~\mu \text{m}$ in diameter. Indeed, while LA-VCSELs are desirable as they provide more optical power, they are heavily multimode. This is quite advantageous for our purposes, but makes them undesirable for industrial applications such as camera auto-focus optics or LiDAR pointers \cite{chang1990transverse}.  As a result, the industry has shifted towards arrays of single mode VCSELs rather than engineering bigger VCSELs.

\begin{figure}[h]
\begin{center}
\includegraphics[scale = 0.65]{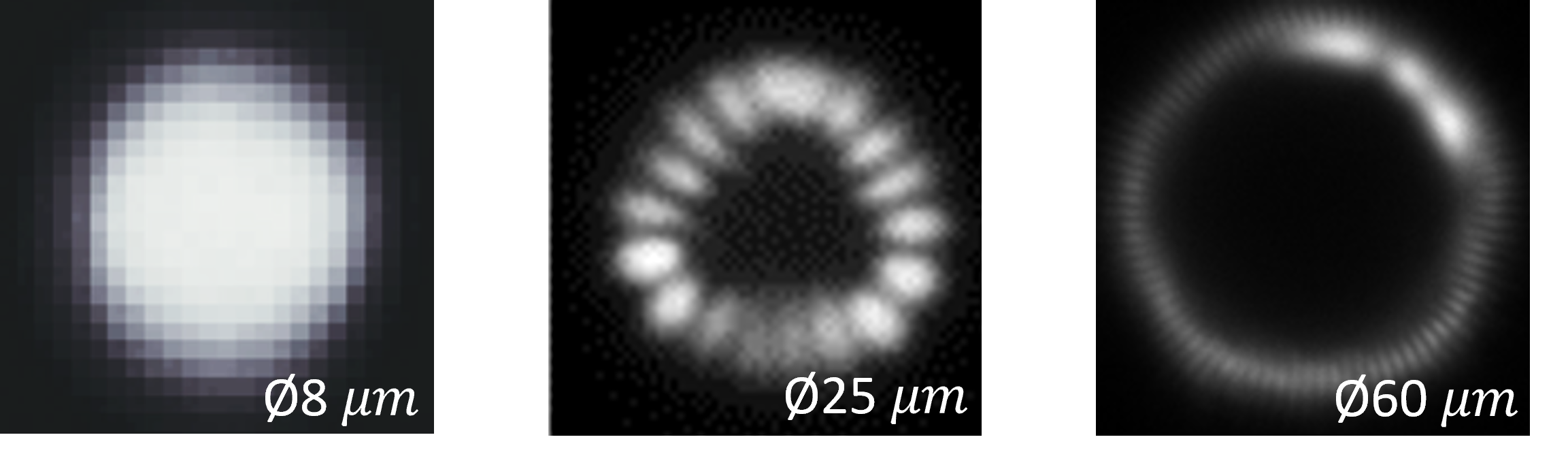}  
\caption{Emission mode profile for 3 different VCSELS with oxide aperture diameters ($\phi$) of, respectively, ${\sim} 8$, $20$, and $60\ \mathbf{\mu}m$.}
\label{fig:mode_round_size}
\end{center}
\end{figure}

Figure \ref{fig:mode_round_size} shows VCSEL emission mode profiles for 3 VCSELs of different emitting diameters, respectively with apertures of $8$, $20$, and $60~\mathbf{\mu}$m. The smallest device shows a single mode Gaussian intensity distribution, while the bigger devices show increasingly high order ring-like mode profiles, also referred to as daisy modes\cite{degen1999transverse}.\par
Bigger VCSELs support more optical modes, should we therefore just use or design the biggest VCSEL possible? Unfortunately, the answer is not obvious, as small devices have a higher concentration of charge carriers close to the center of the cavity. However, as the diameter of the device increases, the emission profile becomes exceedingly ring-like and the center of the device exhibits a drop in power density. Indeed, carriers are injected into the device's active region, the quantum well, by the electrical contacts located at the rim around the VCSEL's aperture. Unfortunately, carrier diffusion is limited to ${\sim}$10~$\mu$m \cite{xiong1998current}, and increasing the device's diameter beyond this distance results in a diminishing carrier concentration towards the device's center. As a result, high order ring modes have access to most of the device's gain.

As a consequence, with substantially bigger devices, most of the LA-VCSEL's central area is dark. The result is an increasingly faction of the device area that cannot achieve lasing, which consequently results in weaker input-output nonlinear transformation.  Therefore, for this type of cavity design, there is a tradeoff between the total area of the VCSEL and its lasing or 'active' area. Furthermore, for the same reason, bigger devices have higher threshold currents, and in our experience, devices exceeding diameters of $100\ \mathbf{\mu}m$ and higher in diameter were not able to lase because before crossing the lasing threshold, the thermal rollover \cite{baveja2011assessment} resulted in a too-strong reduction in optical gain. How then can we make more efficient use of an LA-VCSEL's area, i.e. turn more of the VCSEL area active? The so called 'chaotic cavity VCSEL" design, addresses this point \cite{alkhazragi2023modifying,alkhazragi2023chaotic}.

\subsubsection{Chaotic cavity VCSELs}

Historically, chaotic cavities were studied and proposed as a solution to for example, provide more directional emission from the cavity, especially in the context of micro-lasers \cite{song2009chaotic,shinohara2009ray}.
In the context of LA-VCSELs, chaotic cavities, i.e. cavities breaking their circular symmetry, have been used primarily as a way to achieve a more uniformly active mode profile, a higher density of supported optical states \cite{brejnak2021boosting}, as well as reduction in the mutual coherence between these modes. The reduced coherence in particular is interesting for applications in speckle-free imaging and free-space optical communications \cite{alkhazragi2023modifying,alkhazragi2023chaotic}.\par

In our case, by breaking the circular symmetry of the cold cavity, the high order ring modes are no longer supported for certain designs, which in turn means that optical gain can be "used" by other modes which have a more distributed profile. In practice, this means that a higher fraction of the VCSEL's area is active and supports lasing emission. In turn, this should in principle increase the number of neurons we can use. In \cite{alkhazragi2023modifying,alkhazragi2023chaotic}, the authors report an increased fraction of the device's lasing surface and an increase in the number of modes when switching to a chaotic LA-VCSEL cavity design. In addition, in \cite{brejnak2021boosting} the authors report an increase in emitted power and in quantum efficiency when switching to a chaotic cavity design. In addition the more uniform density of optical modes creates less modal competition, which in turn helps the overall efficiency of the device as the recombination of carriers through non-radiative processes is reduced \cite{brejnak2021boosting}.\par

\begin{figure}[h!]
    \begin{center}
    \includegraphics[scale = 0.5]{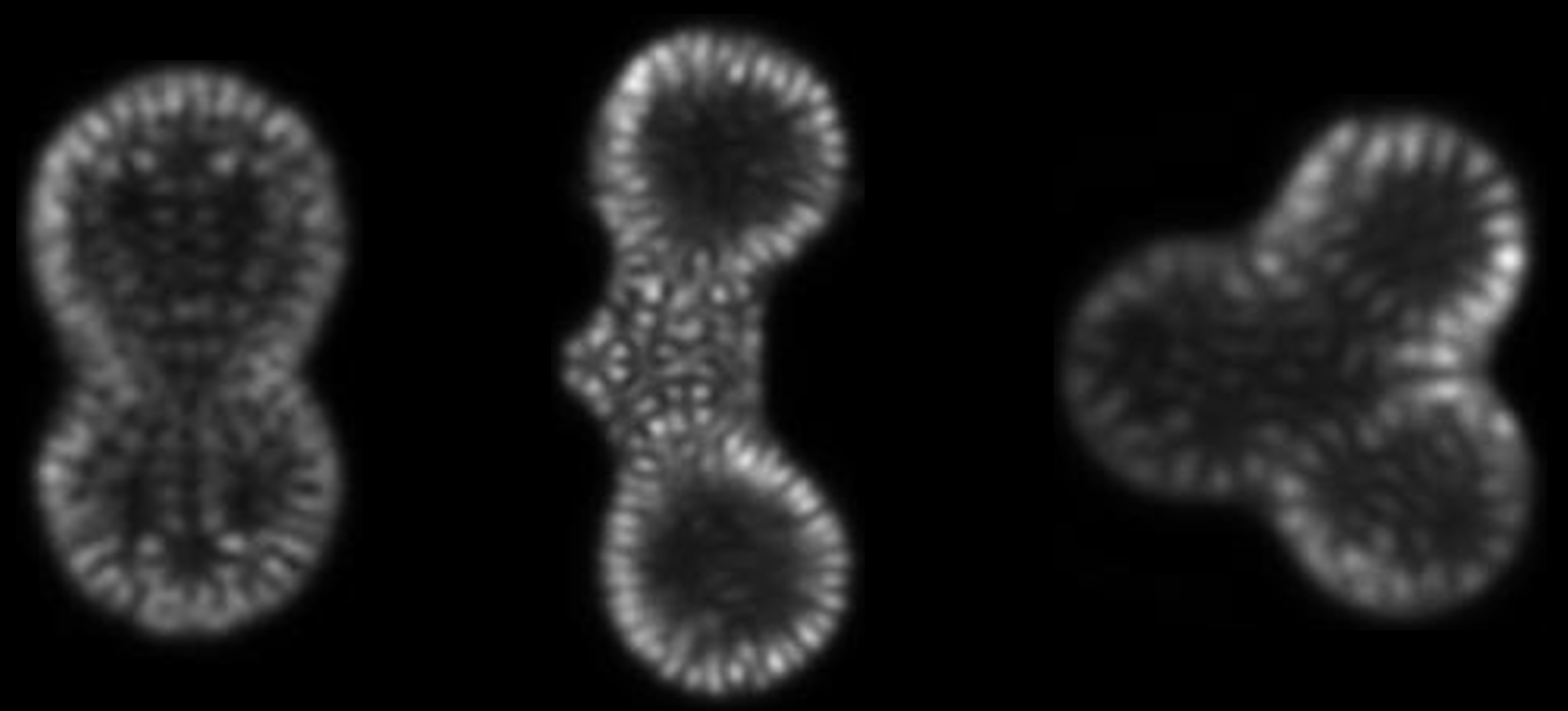}  
    \caption{Chaotic cavity VCSELs with different geometries, showcasing a more active mode profile with less non-lasing regions as compared to the circular devices in Fig. \ref{fig:mode_round_size}.}
    \label{fig:mode_chaotic}
    \end{center}
    \end{figure}

Figure \ref{fig:mode_chaotic} shows the emission mode profile of several LA-VCSELs with non-circular symmetric geometries, showcasing a more active mode profile with less dark regions. The study of the cavity geometry's influence on computational performance remains an extremely interesting and so far unexplored research direction. In \cite{kim2023impact} the authors discuss possibilities of tuning the dynamical response of microlasers by adjusting the cavity geometry, with a potential avenue for chaos-based applications leveraging compact microlasers. Potential applications for these devices are the development of high-power broad-area lasers with stable dynamics and compact lasers for chaos-based applications \cite{brejnak2021boosting}.

\subsection{Updated experimental setup}

\begin{figure}[h!]
    \begin{center}
    \includegraphics[width=1\linewidth]{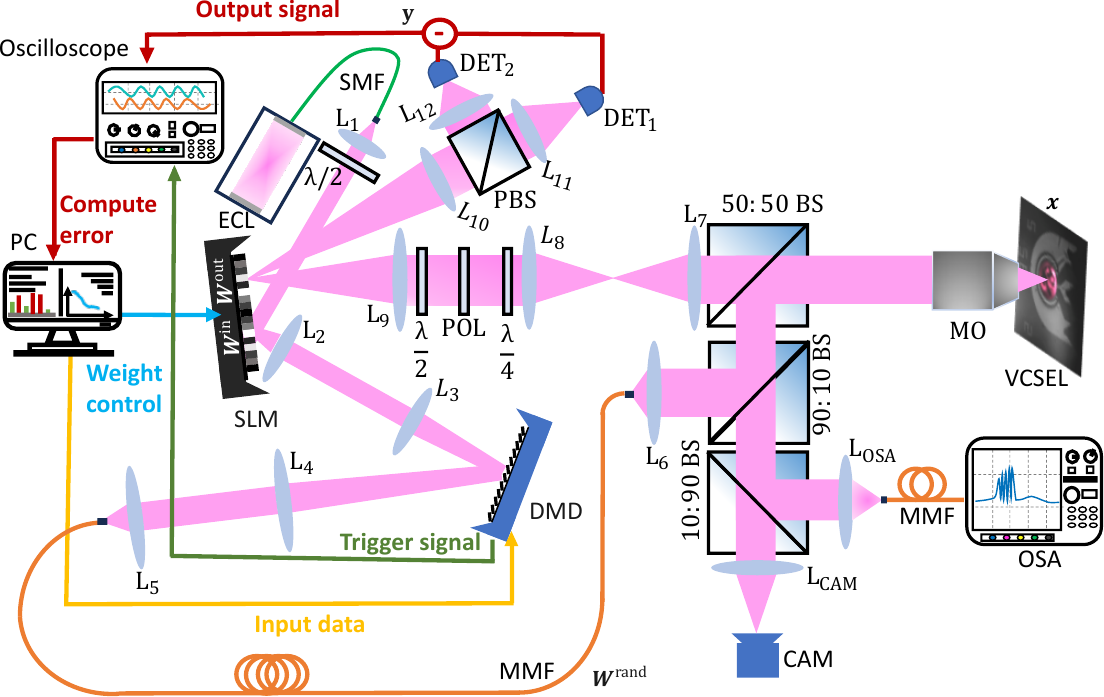}
    \caption{Updated ONN setup. The ONN is now fully tunable and no longer is a reservoir, a single SLM is used to provide, both, input and output weights using, respectively, phase and itensity modulation via a PBS.}
    \label{fig:final_setup}
    \end{center}
    \end{figure} 

The experimental setup presented in Fig. \ref{fig:final_setup} is an updated version of the first experimental setup presented in \cite{porte2021complete,skalli2022computational,skalli2025annealing}. Taking into account the results of the ceiling analysis presented in section \ref{sec:ceiling}, we have updated our experimental setup to include several new features. An SLM was added before the input DMD to provide tunable input weights, while the output DMD was replaced by the same SLM. The SLM screen is large enough to accommodate two beams located on distinct spatial positions to prevent crosstalk. The first corresponding to the input laser beam, the second to the LA-VCSEL. The input weights apply up to $2\pi$ phase modulation to the input data displayed on the DMD before mixing in the MMF, therefore providing positive and negative input weights. Output weights are realized using an intensity modulation configuration, using a polarizing beamsplitter (PBS) and two detectors on each of the PBS-outputs. The detectors' outputs are subtracted electronically in real-time, which in turn provides positive and negative output weights.\par

\begin{figure}[h!]
    \begin{center}
    \includegraphics[width=1\linewidth]{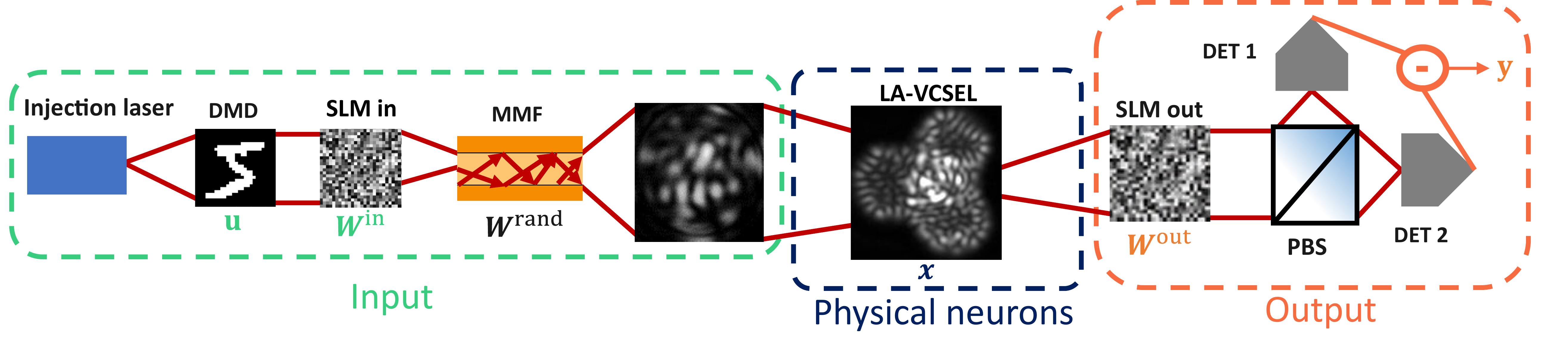}
    \caption{Simplified experimental scheme of the fully tunable ONN.}
    \label{fig:ONN_final_scheme}
    \end{center}
\end{figure}

The number of pixels used on the SLM is determined by the size of the input beam and the LA-VCSEL image on its surface, respectively, and their position is determined by a knife edge measurement while the SLM is in intensity modulation. Finally, using an SLM (Santec LCOS-SLM200) provides up to 10-bit resolution on the weights. The following subsections will discuss the implementation of these improvements in more detail. A simplified diagram illustrating the scheme of our fully tunable ONN is presented in Fig. \ref{fig:ONN_final_scheme}.

\subsubsection{Input layer}

\begin{figure}[h!]
    \begin{center}
    \includegraphics[width=1\linewidth]{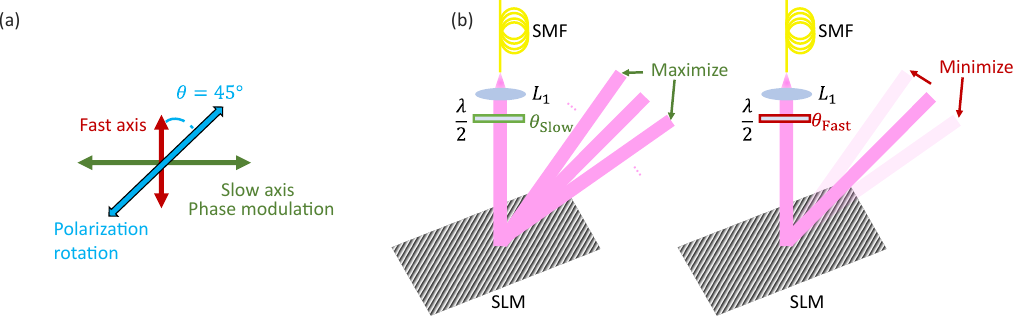}
    \caption{(a) Schematic of the the SLM's refractive index-ellipsoid. (b) SLM displaying a blazed grating, the input polarization is rotated using a half-waveplate to maximize the intensity of high difraction orders to ensure phase-only modulation.}
    \label{fig:SLM_phase_mod}
    \end{center}
    \end{figure}

In our implementation of optical input weights, a collimated input beam was reflected off the SLM's surface onto the DMD. The modulated input data was then coupled into the MMF via imaging. To ensure we are fully modulating the phase of the input beam, we need to align the input beam's polarization to the SLM's slow axis. This can be achieved by displaying a blazed grating on the SLM and rotating the input polarization using a half-waveplate to maximize the intensity of diffraction orders, as shown in Fig. \ref{fig:SLM_phase_mod}.

Via a knife edge measurement, we determine the number of SLM pixels, i.e. input weights that can be used on the input beam. Furthermore, we can adjust the number of input weights by defining a software weight matrix of an arbitrary size smaller than the number of illuminated SLM pixels and resizing it via interpolation to fit the area corresponding to the input beam. In this context, each weight corresponds to a super pixel on the surface of the SLM. 

Yet, because of their size of $8~\mu\text{m}$, the SLM pixels cause diffraction, which in turn creates a loss of coupling efficiency at the MMF due to high NA components of the reflected beam. This ultimately results in a loss of information and reduces the injected power ratio and includes a weight-dependent artificial loss, which is detrimental in terms of performance.

\begin{figure}[h!]
    \begin{center}
    \includegraphics[width=1\linewidth]{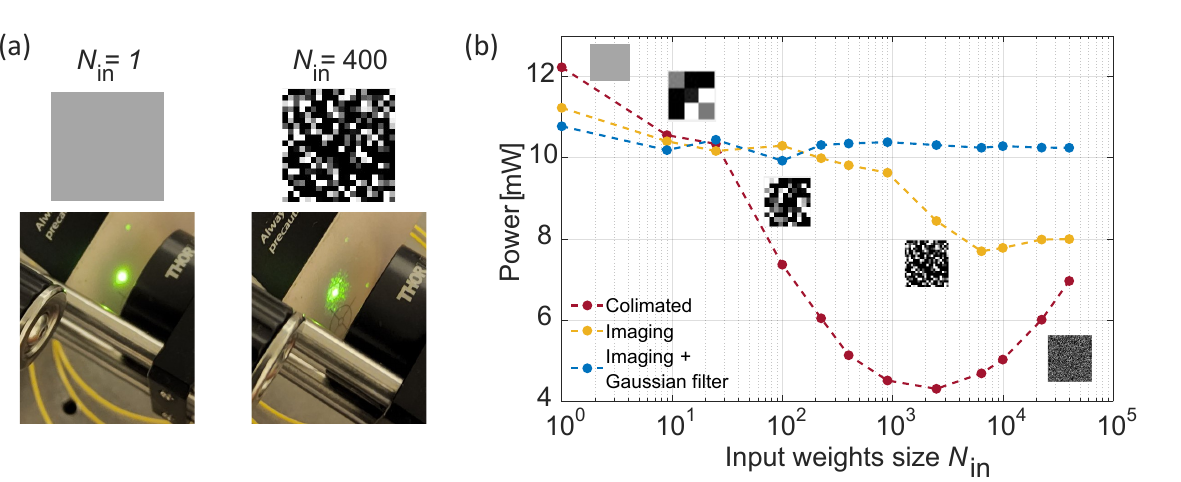}
    \caption{(a) Picture showing the effect of diffraction for larger input weight matrices. (b) Power coupled into the MMF as a function of the input weight matrix size.}
    \label{fig:input_weight_res}
    \end{center}
    \end{figure}

Figure \ref{fig:input_weight_res} shows the effect of diffraction on the input beam. The number of input weights $N_{\text{in}}$ is progressively increased, while the power coupled into the fiber is continuously monitored. As $N_{\text{in}}$ grows, the size of super pixels decreases, which in turn increases the effective NA incident on the MMF, which reduces the coupling efficiency, as when the beam's NA starts exceeding the MMF's NA or when the resulting collimated beam diameter exceeds the imaging lens's diameter, see red curve in Fig. \ref{fig:input_weight_res}(b). After a certain point, the power coupled into the MMF starts to increase again, this is due to an increased spatial frequency that is so high that the phase is no longer modulated, i.e. there is a drop in diffraction efficiency linked to high spatial frequency.
Figure \ref{fig:input_weight_res} (a) shows a picture of the beam before $L_5$ in the case where $N_{\text{in}} = 1$ and $N_{\text{in}} = 400$ respectively, and the effect of diffraction is quite clear; a higher NA translates into a bigger beam diameter in collimated space in between $L_4$ and $ L_5$. 
To mitigate this effect, we decided to image the surface of the SLM onto the DMD using a 2-lens Relay-lens system. This means that the beam no longer propagates in collimated space from the SLM to the DMD, reducing diffraction-induced losses between lenses due to eventual beam clipping. This greatly alleviate the power loss at the MMF coupling stage, as shown by the yellow curve on panel (b).\par

Furthermore, knowing the lenses in our system we can estimate the smallest feature size on the SLM that will couple into the MMF before the corresponding NA will exceed the MMF's NA. Knowing this characteristic feature size, we can apply Gaussian filtering to the input weights to prevent features from going below said size, which yields the blue curve on Fig. \ref{fig:input_weight_res}(b). To estimate the characteristic feature size, we use the following simple approximations: 

\begin{equation}
    {\text{NA}_{\text{DMD}}} = \frac{f_2}{f_3} * {\text{NA}_{\text{SLM}}},
\end{equation}

\begin{equation}
    {\text{NA}_{\text{MMF}}} = \frac{f_4}{f_5} * {\text{NA}_{\text{DMD}}},
\end{equation}

\begin{equation}
    {\text{NA}_{\text{MMF}}} = \frac{f_2*f_4}{f_3*f_5} * {\text{NA}_{\text{SLM}}}.
\end{equation}

Next, we can estimate the diffraction limit of the SLM's characteristic feature size $x_{\text{SLM}}$ by using the Rayleigh criterion:

\begin{equation}
    {\text{NA}_{\text{SLM}}} = \frac{\lambda}{x_{\text{SLM}}},
\end{equation}

Where $\lambda \simeq 975~ \text{nm}$ is the wavelength of the input laser. Combining previous equations yields:

\begin{equation}
    x_{\text{SLM}} = \frac{\lambda*f_2*f_4}{\text{NA}_{\text{MMF}}*f_3*f_5}.
\end{equation}

Finally, using $8~ \mu\text{m}$ as the SLM's pixel pitch, we get the number of pixels that have to be used to define our Gaussian filter kernel. Using $\lambda \simeq 975 ~\text{nm}$, $f_2 = 100 ~\text{mm}$, $f_3 = 100 ~\text{mm}$, $f_4 = 250 \text{mm}$, $f_5 = 20 ~\text{mm}$, $\text{NA}_{\text{MMF}} = 0.22$, we find that the Gaussian kernel should be at least ${\sim} 2$ pixels wide to ensure that the smallest feature size on the SLM still couples to the MMF. We use a Gaussian kernel of width $4$ to filter the input weights. While operating the setup, input weights are rescaled to the interval $[0, 1023]$ to match the 10-bit resolution $2\pi$ modulation of the SLM.

\subsubsection{Output layer}

The modifications to the output layer were quite significant. We now use intensity modulation, and two detectors, one for positive, the other for negative weights. First, to ensure we are in intensity modulation, the incident LA-VCSEL light needs to be linearly polarized and aligned at 45 degrees with respect to the slow axis. To achieve this, the LA-VCSEL goes through a quarter waveplate and linear polarizer, which are then adjusted to maximize the intensity transmitted through the linear polarizer. Subsequently, a half waveplate is rotated to align the linearly polarized LA-VCSEL emission to 45 degrees with the SLM slow axis. This means that the SLM is effectively doing polarization rotation of the LA-VCSEL light. Finally, a PBS acts as a second polarizer to achieve intensity modulation. The two arms of the PBS are polarized perpendicularly to each other, and the output of the detectors $\text{DET}_1$ and $\text{DET}_2$ are subtracted electronically in real-time to achieve positive and negative weights.

\begin{figure}[h!]
    \begin{center}
    \includegraphics[width=1\linewidth]{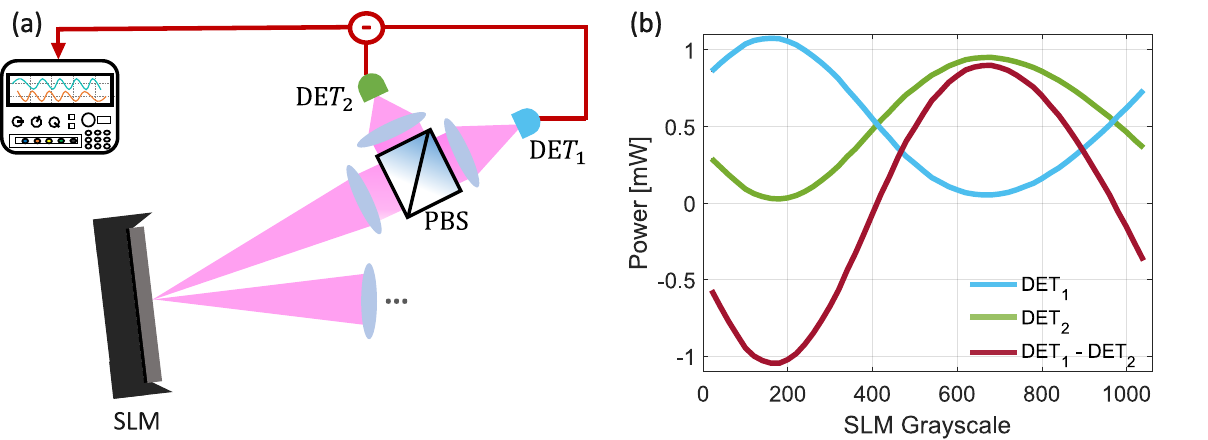}
    \caption{(a) ONN output layer using a SLM, a PBS and two photodetectors. (b) Optical power on $\text{DET}_1$ and $\text{DET}_2$ as a function of the SLM grayscale showing $2\pi$ intensity modulation. The signals from $\text{DET}_1$ and $\text{DET}_2$ are subtracted to yield positive and negative output weights.}
    \label{fig:output_weights_PBS}
    \end{center}
    \end{figure}

Figure \ref{fig:output_weights_PBS} shows how we achieve positive and negative weights. Output weights in the interval $[-1,1]$ are rescaled to the interval corresponding to the range between the minimum and maximum of intensity modulation the SLM provides i.e. $[{\sim} 170, {\sim} 660]$. Finally, the position of the output weights is determined via a knife edge measurement as with one of the two photodetectors to ensure that input and output weights do not overlap.

\section{Benchmarking Training Strategies}
\label{sec:hardware_HP_scan}
After updating the experimental setup to include input and output weights, we now conduct an analysis of the training strategies presented in section \ref{sec:strategies}. We study the impact of the main hyperparameters within each optimization algorithm on performance. Due to the high number of hyperparameters, we use "one-vs-all" classification on the hardest digit of the MNIST dataset, which is digit eight. The main limiting factor in this regard is the slow response time of the SLM of ${\sim} 3~\text{Hz}$ and the scalar output given by the single photodetector. In conjunction, these two factors make hyperparameter tuning a time-consuming process that is impractical on the full MNIST dataset. This section aims to provide guidance for the hyperparameters in each optimization algorithm and gives valid ranges for good performance in a given context. We then use the best hyperparameters for each strategy to compare them in terms of performance.

\subsection{Hyperparameter tuning of training algorithms}

In this section, we solely focus on finding appropriate hyperparameter ranges for all training strategies. Due to hyperparameter scans taking a significant amount of time, numbers presented here are not comparable across different algorithms. The next section focuses on a direct comparison of relevant algorithms using the best hyperparameters found in our analysis. All hyperparameter dependencies were conducted under optimal injection locking conditions, as presented in \cite{skalli2022computational}.

\subsubsection{Finite difference}

\begin{figure}[h!]
    \begin{center}
    \includegraphics[width=1\linewidth]{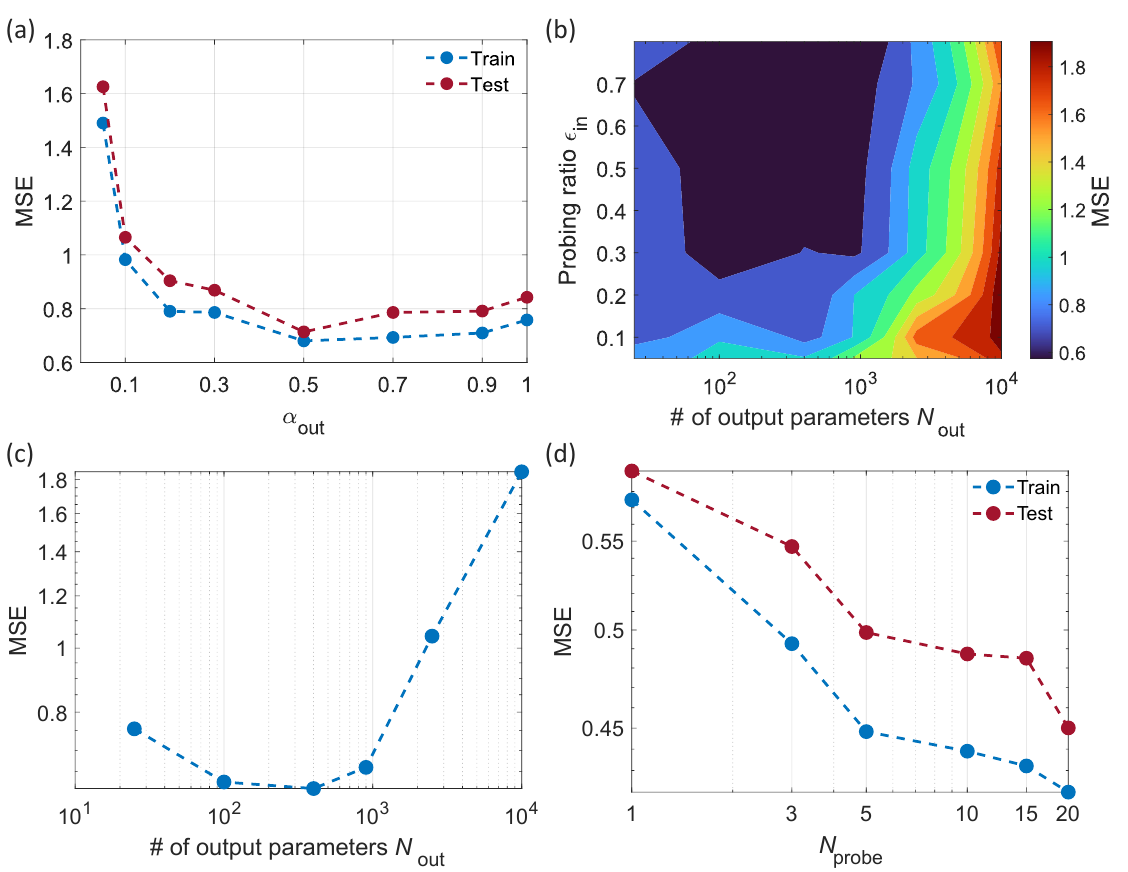}
    \caption{Hyperparameter scan for the output weights using the finite difference method. (a) MSE as a function of the learning rate $\alpha_{\text{out}}$. (b) Colormap of the MSE as a function of $\epsilon$ and $N_{\text{out}}$.  (c) MSE as a function of $N_{\text{out}}$. (d) MSE as a function of $N_{\text{probe}}$.}
    \label{fig:FD_hyperparams}
    \end{center}
    \end{figure}

The finite difference (FD) method is the easiest to implement conceptually, nevertheless, in-hardware learning adds a significant layer of complexity. Crucially, as a result of hardware resolution and noise, infinitesimal values of $\epsilon$ would not yield any change in the output as measured by the oscilloscope with an 8-bit resolution when using its full range.
This effectively sets a lower bound on $\epsilon$. Moreover, the optimal value of $\epsilon$ should, in principle, depend on the size of the super pixels on the SLM representing the weights. Indeed, the bigger the superpixel, the more optical intensity it receives, which makes it easier to detect changes at the output when perturbing said weight. Therefore $\epsilon$ should increase as the number of superpixel increases (i.e. size of superpixels decreases). First, let us focus on the output parameters, and Fig. \ref{fig:FD_hyperparams} shows the results of the hyperparameter scan for the output weights.

First, the influence of the learning rate $\alpha_{\text{out}}$ is studied in Fig. \ref{fig:FD_hyperparams}(a), showing that $\alpha_{\text{out}} = 0.5$ is an optimal value. Then, we show how $\epsilon$ changes with the size of the output weight matrix in Fig. \ref{fig:FD_hyperparams}(b). Interestingly, despite what we might assume, as the size of the output weight matrix increases, the optimal values of $\epsilon$ remain unchanged, our initial assumption is therefore false. In addition, optimal values of $\epsilon$ are quite high at around ${\sim} 0.5$, while output weights are in the interval $[-1,1]$, this highlights the rather counterintuitive nature of hardware learning as such a high $\epsilon$ value would be absurd in the context of a software NN. Furthermore, performance degrades significantly when increasing the number of output weights. 
As the number of superpixels increase, the dimensionality of the search problem also increases, as a consequence the FD methods should need more iterations to reach the same level of performance. Consequently, for a fixed training time, increasing the dimensionality of the search space i.e. number of output weights results in a significant decrease in performance, showing that the optimal size of the output weight matrix for the FD method is only $N_{\text{out}} = 400$ as shown in Fig. \ref{fig:FD_hyperparams}(c). This constitutes a major limitation of the FD algorithm in the context of our ONN and severely diminishes its applicability.

\begin{figure}[h!]
    \begin{center}
    \includegraphics[width=1\linewidth]{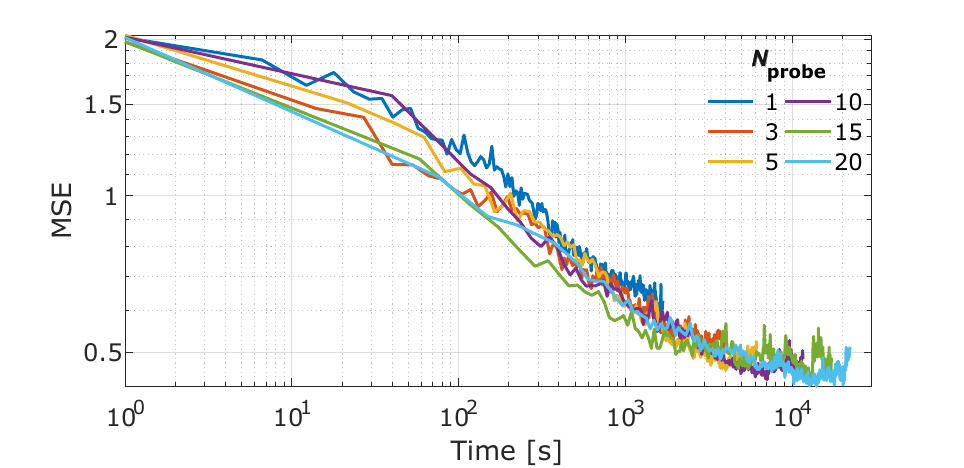}
    \caption{Performance of the FD method as a function of training time for different values of $N_{\text{probe}}$.}
    \label{fig:FD_time}
    \end{center}
    \end{figure}

Next, we studied the impact of the number of probed weights per epoch $N_{\text{probe}}$. Here $N_{\text{probe}}$ weights are each perturbed sequentially, i.e. one at a time, to build a gradient vector and they are then updated at the same time. In terms of scaling this means that if $N_{\text{probe}} = 20$, each epoch takes 20 times longer, as 20 superpixels are perturbed one after the other. Figure \ref{fig:FD_hyperparams}(d) shows the impact of $N_{\text{probe}}$ on the MSE for the output weights when training for a fixed number of epochs. We can see that performance increases with $N_{\text{probe}}$, as sampling more coordinates i.e. superpixels yields a more accurate estimate of the gradient vector before updating the weights at every epoch. Yet this comes at the cost of a longer training time. To investigate the influence of $N_{\text{probe}}$ on convergence speed, we study the convergence of the FD method as a function of training time for different values of $N_{\text{probe}}$, as shown in Fig. \ref{fig:FD_time}. Crucially, all curves overlap and converge to the same level showing that $N_{\text{probe}} = 1$ should reach the same performance level as $N_{\text{probe}} = 20$ provided they are both trained for the same amount of time.


We did not study the influence of the FD methods on the input weights, as that would add an additional layer of complexity to the problem, and as shown by our measurements, the FD method is not well suited to high dimensional problems.

\subsubsection{Simultaneous Perturbation Stochastic Approximation}

SPSA can be viewed as a more robust and significantly more efficient variation on the FD method. Moreover, since all weights are perturbed at the same time, we expect $\epsilon$ to be smaller. Finally, we should not in principle assume that input and output weights have the same impact on the loss function minimization and probing. In practice, this would mean that we would have 4 different hyperparameters, $\epsilon_{\text{in}}$, $\alpha_{\text{in}}$, $\epsilon_{\text{out}}$ and $\alpha_{\text{out}}$. For simplicity, we will assume input and output parameters are independent and scan them independently in order to avoid high dimensional hyperparameter grid which can be highly time-consuming.

\begin{figure}[h!]
    \begin{center}
    \includegraphics[width=1\linewidth]{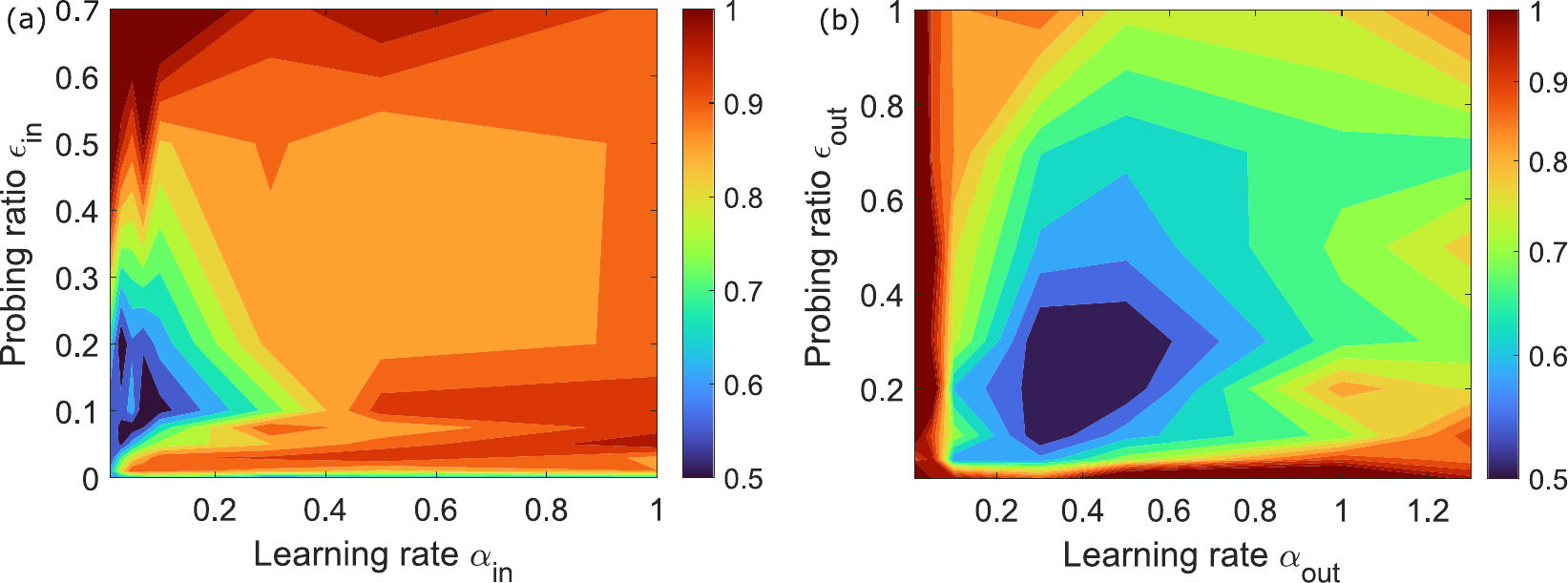}
    \caption{Hyperparameter scan for the input and output weights using the SPSA algorithm. (a) Colormap of the MSE as a function of $\epsilon_{\text{in}}$ and $\alpha_{\text{in}}$. (b) Colormap of the MSE as a function of $\epsilon_{\text{out}}$ and $\alpha_{\text{out}}$.}
    \label{fig:SPSA_hyperparams}
    \end{center}
    \end{figure} 

Figure \ref{fig:SPSA_hyperparams} shows the results of the hyperparameter scan for the input and output weights with the MSE as a colormap. The learning rate $\alpha$ is scanned on the x-axis and the perturbation $\epsilon$ on the y-axis for, both, input and output weights in Fig. \ref{fig:SPSA_hyperparams}(a) and (b), respectively. First, output weights are scanned with an output matrix size of $N_{\text{out}} = 2500$. Optimal values are $\epsilon_{\text{out}} \in [0.2, 0.3]$ and $\alpha_{\text{out}} \in [0.3, 0.4]$. These values are then fixed within the optimal range to $\alpha_{\text{out}}=0.35$ and $\epsilon_{\text{out}} = 0.25$ and $\epsilon_{\text{in}}$, $\alpha_{\text{in}}$, are scanned in the same way, with $N_{\text{in}} = 400$. Crucially, we see that the optimal values for the input hyperparamters are significantly smaller than for the output, with optimal values $\epsilon_{\text{in}} \in [0.08, 0.1]$ and $\alpha_{\text{in}} \in [0.05, 0.1]$, highlighting the complexity of tuning the input weights, i.e. highly non-convex and ragged error landscape. 
Moreover, in our experience, these hyperparameters are quite robust for nearly all values of $N_{\text{in}}$ and $N_{\text{out}}$, and the optimal values are stable even across different LA-VCSELs.

\begin{figure}[h!]
    \begin{center}
    \includegraphics[width=1\linewidth]{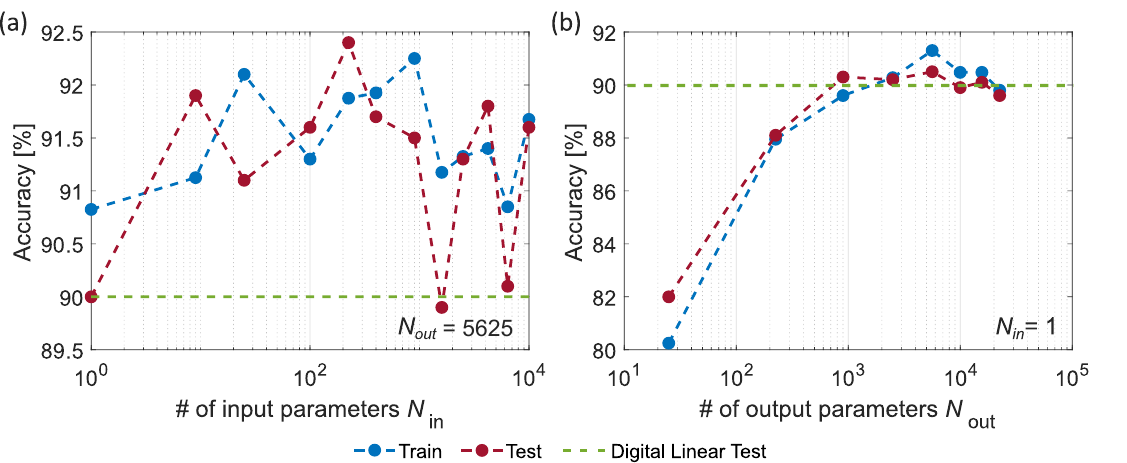}
    \caption{Performance of the SPSA algorithm as a function of the number of (a) input $N_{\text{in}}$  and (b) output $N_{\text{out}}$ parameters.}
    \label{fig:SPSA_params_res}
    \end{center}
    \end{figure}

Finally, contrary to the FD method, SPSA is able to leverage a greater number of parameters and still converge, even for a fixed number of training epochs, making it a simple yet powerful hardware-compatible optimization strategy. As illustrated in Fig. \ref{fig:SPSA_params_res}(b), SPSA is able to converge to a lower MSE than the FD method and performance increases with the number of parameters, and saturates at around $N_{\text{out}} = 2500$ whereas FD saturated at $N_{\text{out}} = 400$. The advantages of SPSA over FD become even more apparent when considering that, in terms of the number of measurements per epoch, SPSA is equivalent to FD with $N_{\text{probe}} = 1$.

While SPSA can be a powerful algorithm, it is not without its limitations. When trying to study the influence of the number of input weights, no clear trend can be extracted as shown in Fig. \ref{fig:SPSA_params_res}(a). It seems that SPSA is not really able to leverage the additional degrees of freedom provided by the input weights.



\subsubsection{Covariance Matrix Adaptation Evolution Strategy}

For the CMA-ES algorithm, the standard deviation along each direction in the search space is learned through sampling. Therefore, we do not need to consider different hyperparameters for input and output weights. The main parameters influencing CMA-ES convergence are the initial standard deviation of the distribution $\sigma_{\text{init}}$ and the population size $p$. In general, CMA-ES is appreciated for its robustness in terms of hyperparameter choice, and many theoretical estimates or rules of thumb for hyperparameter choice as a function of dimensionality exist in the literature \cite{hansen2016cma}. 

\begin{figure}[h!]
    \begin{center}
    \includegraphics[width=1\linewidth]{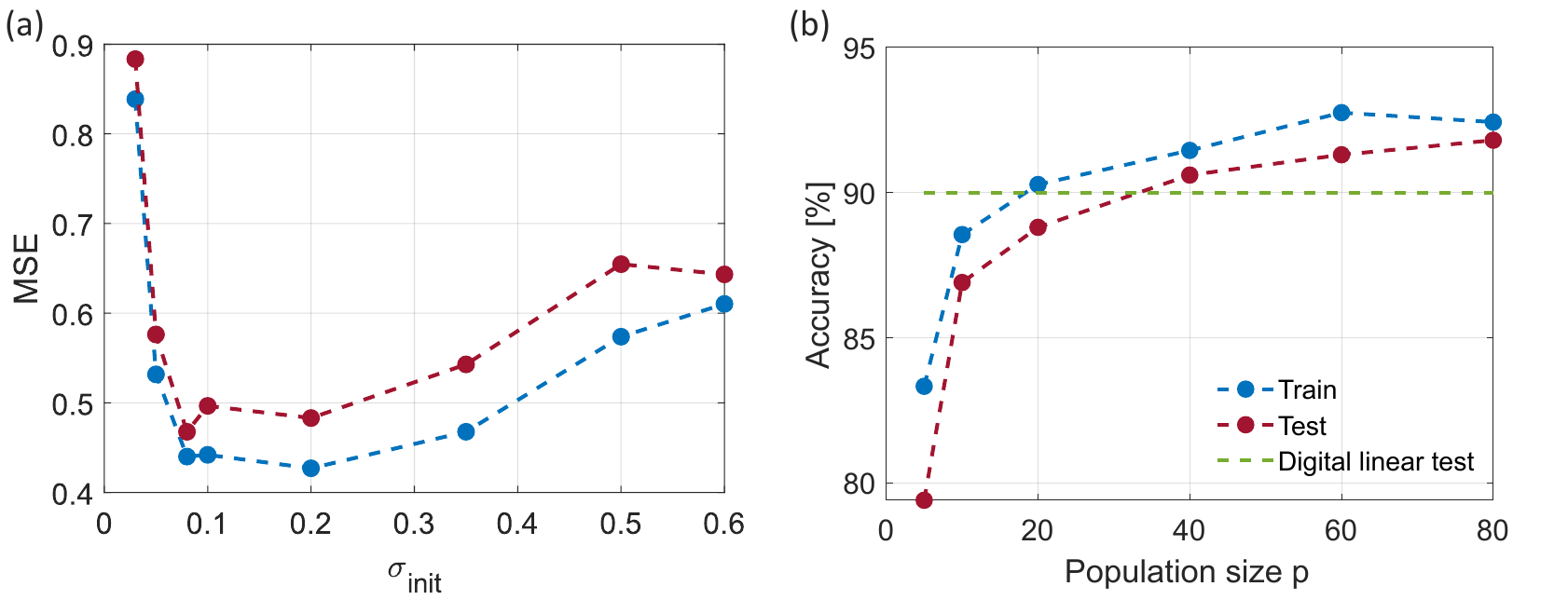}
    \caption{(a) MSE as a function of the initial standard deviation $\sigma_{\text{init}}$. (b) Accuracy as a function of population size $p$.}
    \label{fig:CMA_hyperparam_scan}
    \end{center}
    \end{figure}

Since CMA-ES is an advanced optimization algorithm we study the behavior of these two hyperparameters for the following output and input weight settings respectively; $N_{\text{out}} = 2500$ and $N_{\text{in}} = 2500$ . 
Figure \ref{fig:CMA_hyperparam_scan} shows the impact of $\sigma_{\text{init}}$ on the performance of the CMA-ES algorithm. We see that the optimal value of $\sigma_{\text{init}}$ is quite stable and is around $\sigma_{\text{init}} = 0.2$.

The last and perhaps most impactful parameter on performance is the population size. Performance increases with larger populations and then saturates, which is a general feature of population-based optimization algorithms. For our purposes, the saturation point is at around $p \in [40, 60]$. 

Because of its quadratic scaling with the dimensionality of the optimization problem, tuning the few hyperparameters of CMA-ES can be quite time consuming. As such, we did not explicitly study the performance scaling of CMA-ES with respect to the number of parameters, but we can infer from the previous section that CMA-ES is able to leverage a large number of parameters. Furthermore, we provide in the following section a comparison of all relevant training strategies in terms of performance.

\subsubsection{Parameter-exploring Policy Gradients}

Much like CMA-ES, PEPG has the merit of being quite a robust strategy, with a noticeable advantage being that its computational overhead is much smaller than that of CMA-ES.  
At first glance, it may seem that tuning PEPG hyperparameters is a cumbersome task compared to CMA-ES. Yet in practice, the algorithm is robust and its hyperparameter behavior is quite predictable. Furthermore, its reduced computational overhead allows us to study a greater number of hyperparameters and dependencies in higher dimensional settings. As such we study all hyperparameters with $N_{\text{out}} = N_{\text{in}} = 2500$. \par

\begin{figure}[h!]
\begin{center}
\includegraphics[width=1\linewidth]{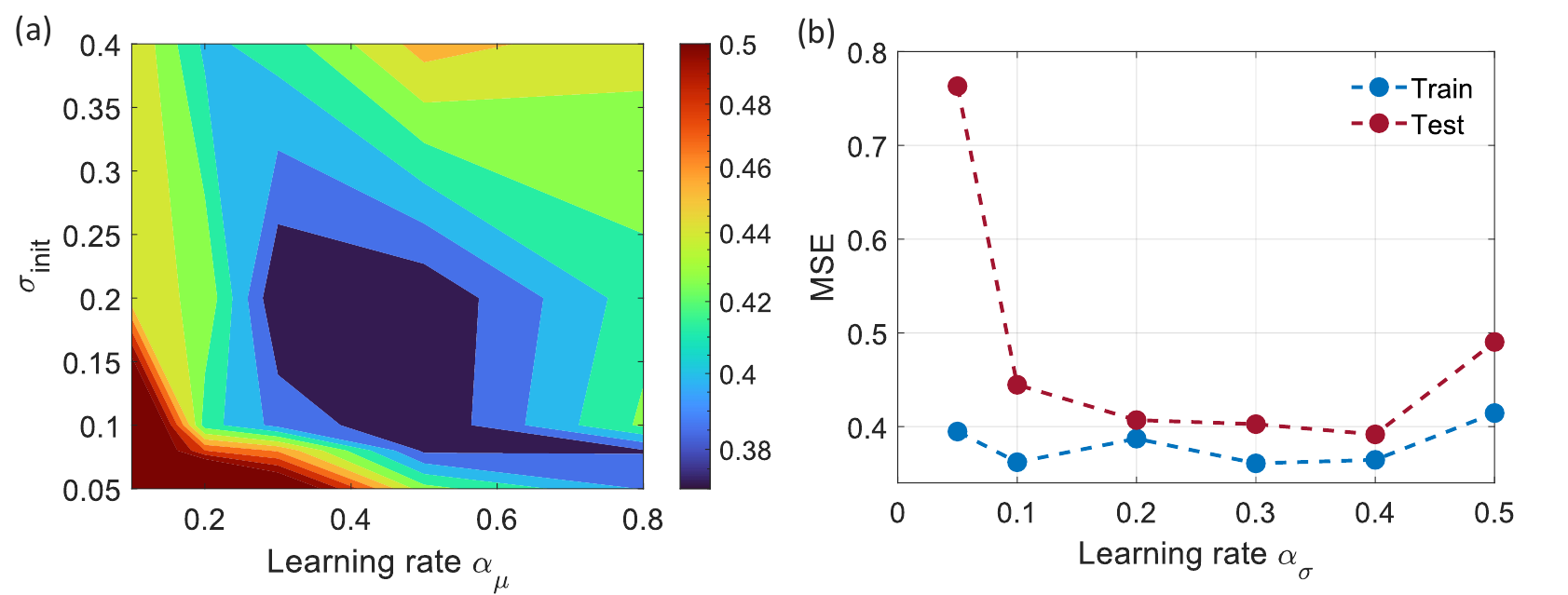}
\caption{Hyperparameter scan for the input and output weights using the PEPG algorithm. (a) Colormap of the MSE as a function of the initial standard deviation $\sigma_{\text{init}}$ and mean learning rate $\alpha_{\mu}$. (b) MSE as a function of the stadard deviation learning rate $\alpha_{\sigma}$. }
\label{fig:PEPG_scan1}
\end{center}
\end{figure}

We start by studying the impact of the initial standard deviation of the normal distribution defining the population $\sigma_{\text{init}}$, as well as the learning rate which adapts the mean $\alpha_{\mu}$. We then study the impact of $\alpha_{\sigma}$, the learning rate which adapts the standard deviation. As shown in Fig. \ref{fig:PEPG_scan1}(a) and (b), respectively. Optimal values are $\sigma_{\text{init}} {\sim} 0.2$ similar to CMA-ES, $\alpha_{\mu} \in [0.4 , 0.5]$ and the choice of $\alpha_{\sigma}$ is not as critical as it shows a rather flat dependency with optimal values ranging between $\alpha_{\sigma} \in [0.2,0.4]$.

\begin{figure}[h!]
\begin{center}
\includegraphics[scale=0.5]{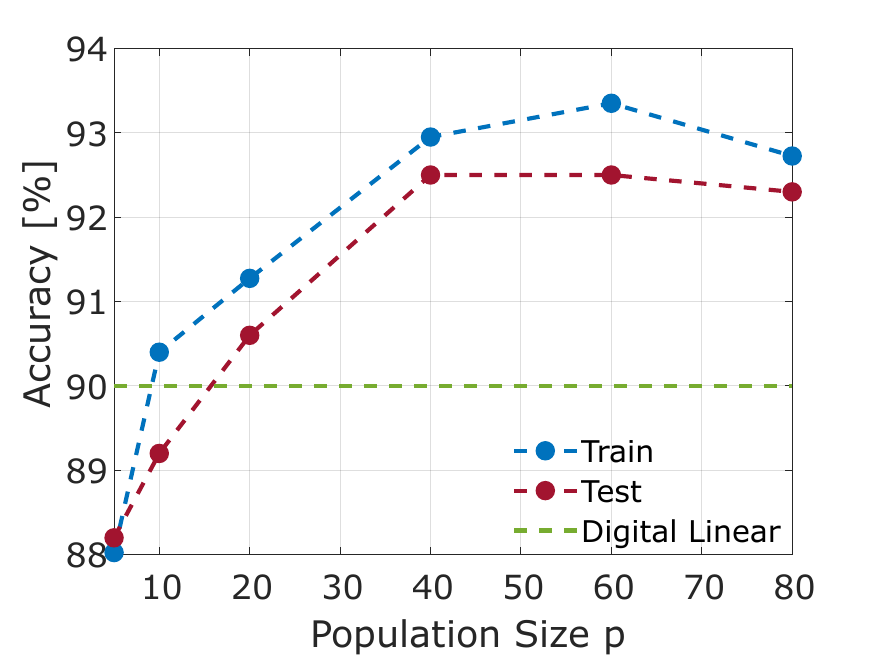}
\caption{Classification accuracy as a function of the population size $p$}
\label{fig:PEPG_scan2}
\end{center}
\end{figure}

We then studied the impact of population size, highlighting the same behavior measured for CMA-ES as shown in Fig. \ref{fig:PEPG_scan2}. For our purposes, the saturation point is at around $p = 40$, which corresponds to ${\sim} 1\%$ of the $5000$ dimensional search space, confirming the same trend we measured on a digital convolutional NN on the MNIST dataset. 

\begin{figure}[h!]
\begin{center}
\includegraphics[width=1\linewidth]{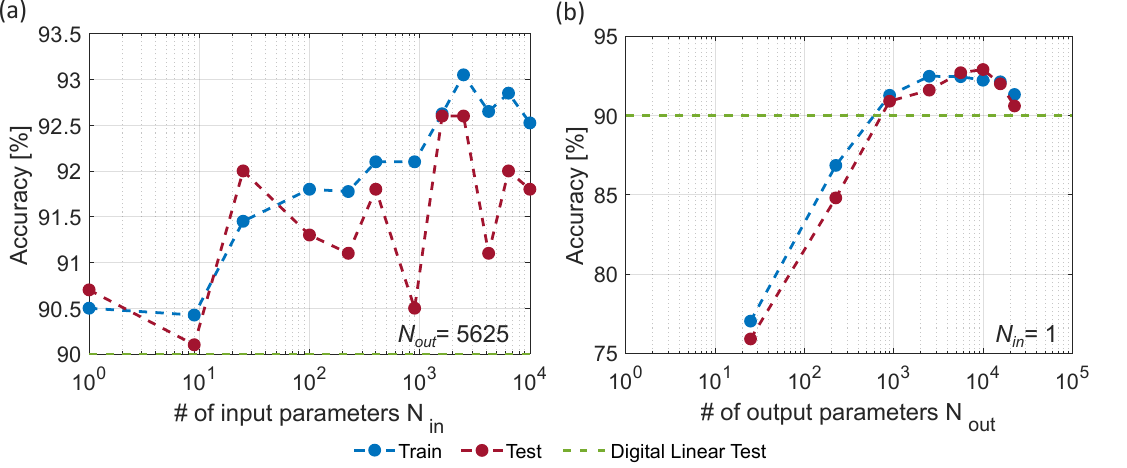}
\caption{Classification accuracy as a function of the number of (a) input $N_{\text{in}}$ and (b) output $N_{\text{out}}$ parameters.}
\label{fig:PEPG_scan3}
\end{center}
\end{figure}

Then, by fixing the population size at $p = 40$ to reduce measurement time, we study how the performance of PEPG scales with the number of input and output parameters respectively. As shown in Fig. \ref{fig:PEPG_scan3}, performance increases with the number of parameters and saturates at around $N_{out} = 5625$ and $N_{\text{in}} = 2500$, showcasing the powerful nature of the PEPG algorithm as opposed to the FD method which saturates at $N_{\text{out}} = 400$, or even SPSA which can leverage more output weights but struggles greatly when it comes to making use of the input weights or reaching similar performance levels. 

\newpage
\section{Comparison of training algorithms}
\label{sec:alg_compare}
We can now compare these different optimization algorithms in terms of convergence efficiency. In this analysis, we specifically focus on SPSA, PEPG and CMA-ES since they show the most promise and can handle high dimensional optimization. Previously, we showed convergence in terms of epochs, which are equivalent to individual optimization steps. Crucially, different algorithms compute these update steps in vastly different ways, and as such, the energy cost of each update or the time spent per epoch is different. By studying convergence as a function of physical time for convergence, we begin to understand the real-world efficiency of each strategy in a hardware context. Indeed, SPSA only takes two measurements per epoch, while CMA-ES and PEPG require a full population evaluation. It would appear therefore that SPSA is more efficient in terms of convergence, but as we will see, the situation is not so trivial. \par

\begin{figure}[h!]
    \begin{center}
    \includegraphics[width=1\linewidth]{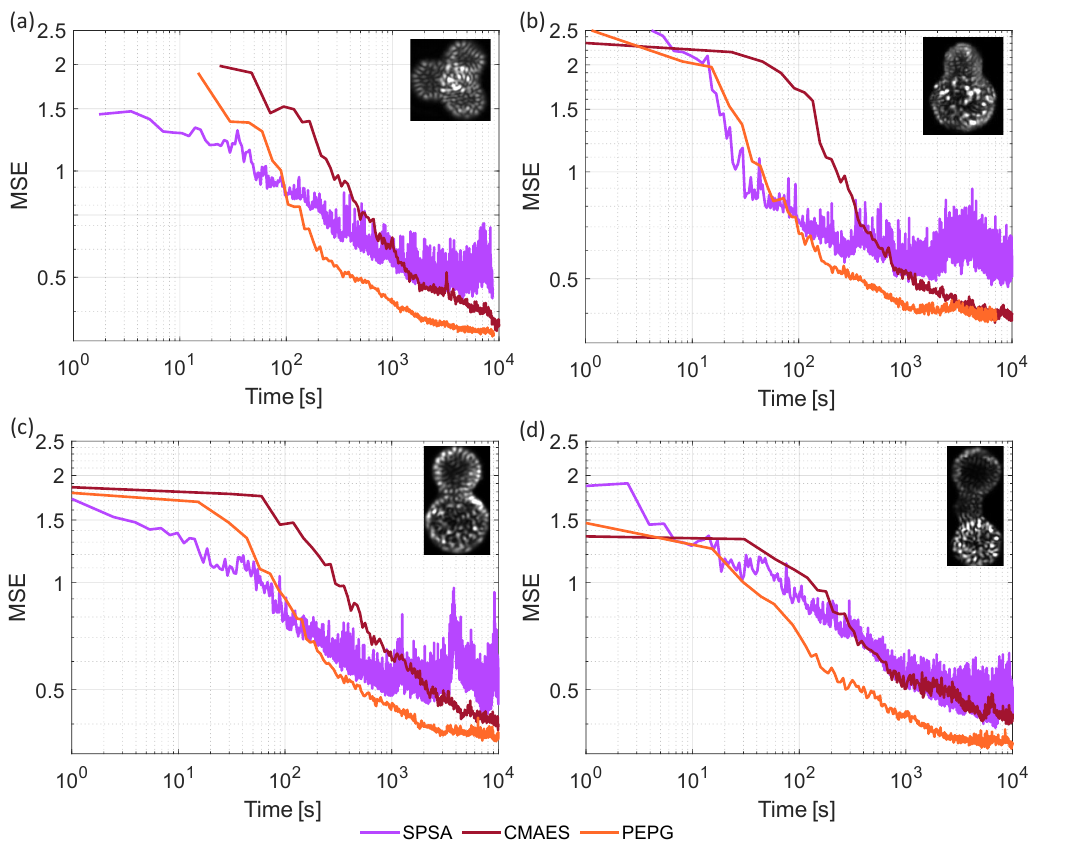}
    \caption{Convergence efficiency for four different chaotic cavity LA-VCSELs trained using SPSA, PEPG and CMA-ES.}
    \label{fig:VCSEL_convergence}
    \end{center}
\end{figure}

Indeed, while SPSA provides quicker updates, these updates are noisier and contain less information than updates computed via PEPG or CMA, making them less efficient in terms of convergence. And yet, one cannot simply neglect the computational overhead incurred by the CMA-ES algorithm, as this overhead scales quadratically with the number of parameters. We extensively compared these algorithms across a wide variety of devices of different shapes and sizes as shown in Fig. \ref{fig:VCSEL_convergence}. As expected, both CMA-ES and PEPG reach better performance levels than SPSA. We comfirmed PEPG to be the most efficient optimization algorithm as it is the most robust and provides the fastest convergence, being anywhere from ${\sim} 3$ to ${\sim} 4$ times faster than CMA-ES in this present test, and reaching significantly better performance than SPSA more efficiently. It should be stressed however that convergence efficiency is highly task dependent.\par 

\begin{figure}[h!]
    \begin{center}
    \includegraphics[width=1\linewidth]{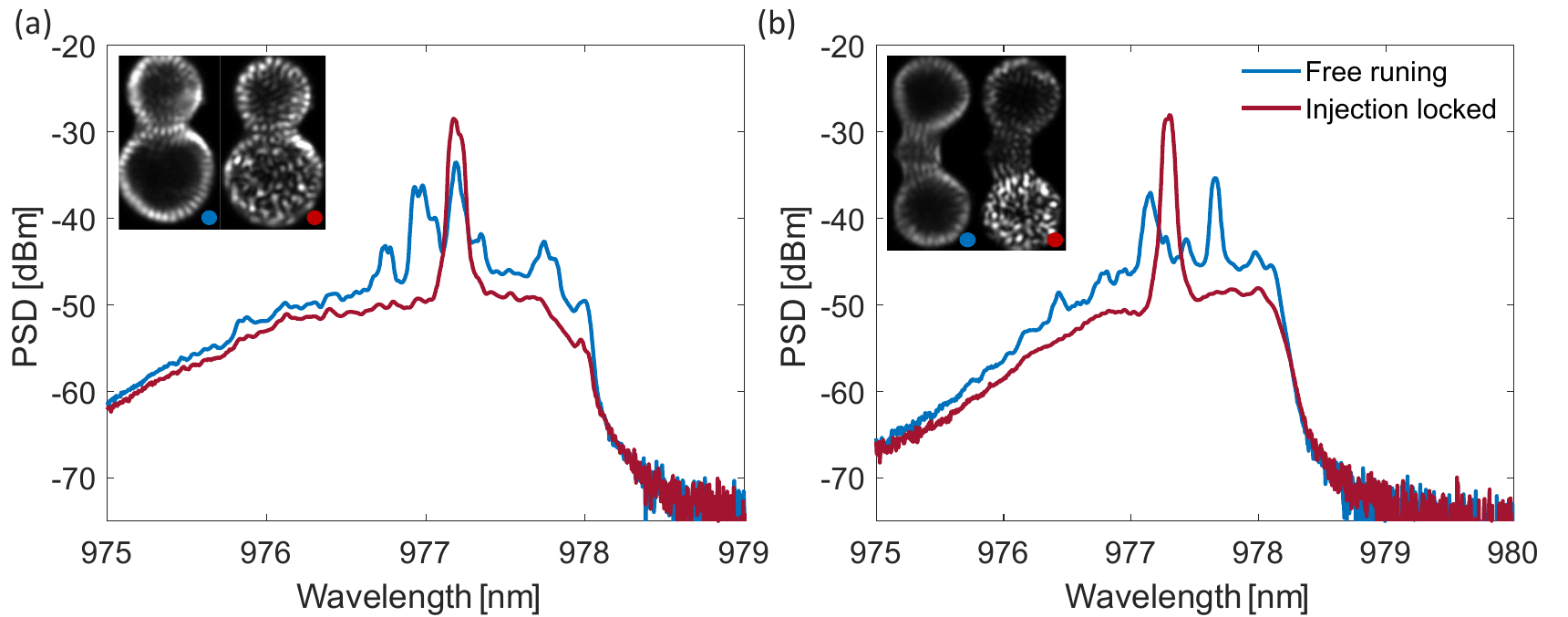}
    \caption{Spectra / mode profiles of two different free running and injection locked chaotic cavity LA-VCSELs.}
    \label{fig:VCSEL_spectra_mode_profile}
    \end{center}
\end{figure}

For reference, our performance evaluation was conducted under optimal injection locking conditions to ensure maximum performance as presented in \cite{skalli2022computational}, the free running and locked mode profile as well as spectra are shown in figure \ref{fig:VCSEL_spectra_mode_profile} for two selected LA-VCSELs.\par

\begin{figure}[h!]
    \begin{center}
    \includegraphics[width=1\linewidth]{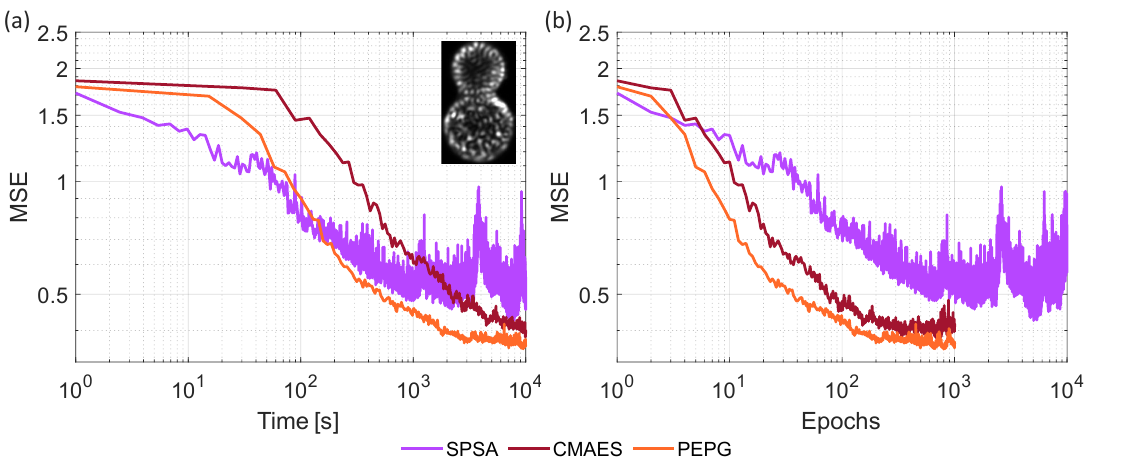}
    \caption{Convergence of a selected LA-VCSEL in terms of time (a) and epochs (b) for the different optimization algorithms.}
    \label{fig:VCSEL_convergence_epochs}
    \end{center}
\end{figure}

To illustrate how misleading a comparison in terms of epochs can be, we show the convergence of a selected LA-VCSEL, both, in terms of time and epochs in Fig. \ref{fig:VCSEL_convergence_epochs}(a) and (b), respectively. First, if we only consider convergence in terms of epochs, it appears that SPSA is always significantly worse than the other two algorithms. However, if training time is restricted, then SPSA performs better than CMA-ES for this task. Moreover, the difference between PEPG and CMA-ES is greatly diminished when only considering epochs. Indeed, while PEPG takes ${\sim} 140$ epochs to reach an MSE of $0.4$ CMA-ES takes ${\sim} 200$ epochs. This might seem like a small difference, yet if we look at the time spent to reach each value, PEPG takes ${\sim} 1900$ s while CMA-ES takes ${\sim}6240$ s making PEPG ${\sim}3.3$ times faster which cannot be deduced when only considering epochs.  

\begin{figure}[h!]
    \begin{center}
    \includegraphics[width=1\linewidth]{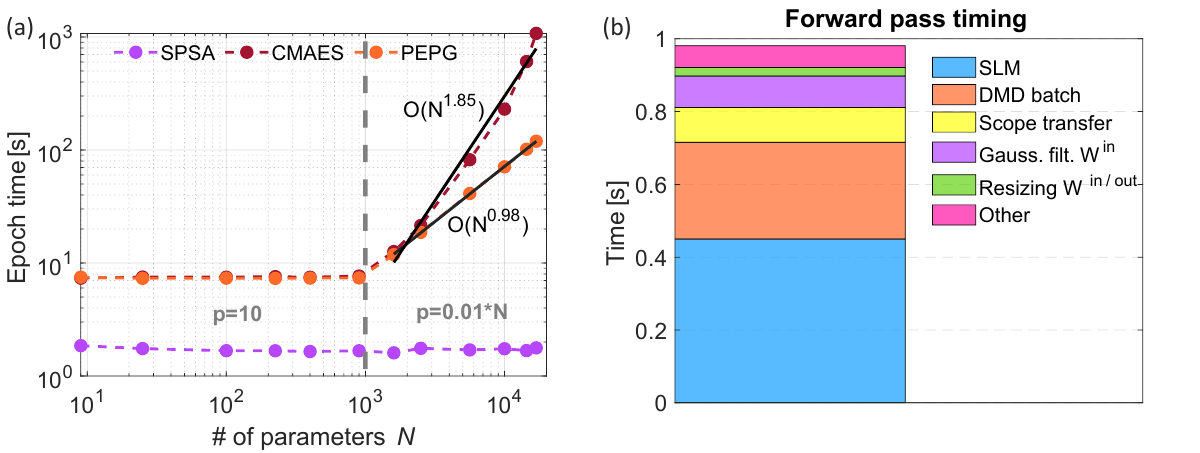}
    \caption{(a) Time taken by each optimization epoch for different training algorithms in seconds. (b) Bar chart detailing the time taken by each step during the forward pass on the experimental setup.}
    \label{fig:VCSEL_timing}
    \end{center}
\end{figure}

To further investigate the behavior and scaling of these different algorithms, Fig. \ref{fig:VCSEL_timing}(a) shows the time spent per epoch for each algorithm as a function of the number of parameters $N$, while (b) shows a detailed breakdown of the time spent at each step of the forward pass for our ONN.

In Fig. \ref{fig:VCSEL_timing}(a), the offset between the population based algorithms and SPSA stems from the larger number of measurements per epoch for PEPG and CMA-ES. Indeed, SPSA keeps a constant epoch time since it only requires two measurements per epoch. In contrast, in Fig. \ref{fig:VCSEL_timing}(a) the population size is either $p=10$ for $N  \leq 1000$, or scales with the dimensionality of the search space according to $p = \lceil0.01 \times N\rceil$, when $N  > 1000$. Crucially, there is a stark difference between PEPG and CMA-ES as the number of parameters increases due to the additional quadratic overhead of CMA-ES induced by the covariance matrix calculation. We were able to experimentally verify that PEPG scales with $O(N^{0.98})$ and CMA-ES with $O(N^{1.85})$. This stark difference reaches close to an order of magnitude for parameter numbers relevant to our experiment i.e. ${\sim} 10000$ parameters. \par 

To partially mitigate this heavy overhead, several variants of the original CMA-ES algorithm have been explored to achieve sub-quadratic scaling, and a detailed comparison for a wide range of synthetic benchmark tasks is provided in \cite{loshchilov2014computationally,varelas2018comparative}. However, it should be stressed however that no such comparison exists in the context of high dimensional NN optimization. This simple characterization is another way of confirming the results shown in Fig. \ref{fig:VCSEL_convergence}. Even if CMA-ES is able to leverage a large number of parameters, its quadratic overhead can automatically disqualify it in the context of optimization under limited digital resources, for all but moderate dimensional physical systems. In addition, we are still comparing these algorithms using a conventional digital computer, Dell Precision i5-6400 8Gb RAM, which means that this difference will be exacerbated when considering limited digital resources such as a Raspberry Pi.  Our comparison of different optimization algorithms and their corresponding overhead is crucial when designing autonomous hardware NN, where the reliance on an external digital computer needs to be minimized. We feel it remains a largely unexplored question in the context of unconventional hardware optimization and deserves more in-depth investigation. \par

Figure \ref{fig:VCSEL_timing}(b), showcases how the SLM we use is our main bottleneck in terms of speed followed by the DMD batch, referring to the total time it takes the DMD to show a batch of 4000 images in this present case. A fundamental limiting factor in our setup is that population samples have to be evaluated sequentially. In contrast, software NNs can benefit from hardware acceleration such as vectorization, and massive parallelization to perform evaluation of population samples in parallel. However, such parallelization only addresses the time-bottleneck, while it is inefficient and counterproductive in terms of energy efficiency \cite{gustafson1988reevaluating,hill2008amdahl}. For the output weights, one could imagine creating copies of the LA-VCSEL and using a detector array to evaluate the population in parallel, yet this would result in a significant increase in experimental complexity and cost. Moreover, multiplexing the input weights in such a way is not possible. At first glance, it would seem that the only immediate avenue for addressing this problem is to increase the speed at which the weights are addressed, i.e. the speed of the SLM, which is currently limited to ${\sim} 3~\text{Hz}$. Crucially, for high dimensional problems, using a faster SLM would only help with the convergence speed of PEPG and SPSA, the CMA-ES algorithm would be largely unaffected due to its computational overhead in a high dimensional optimization scenario.

\begin{figure}[h!]
    \begin{center}
    \includegraphics[width=1\linewidth]{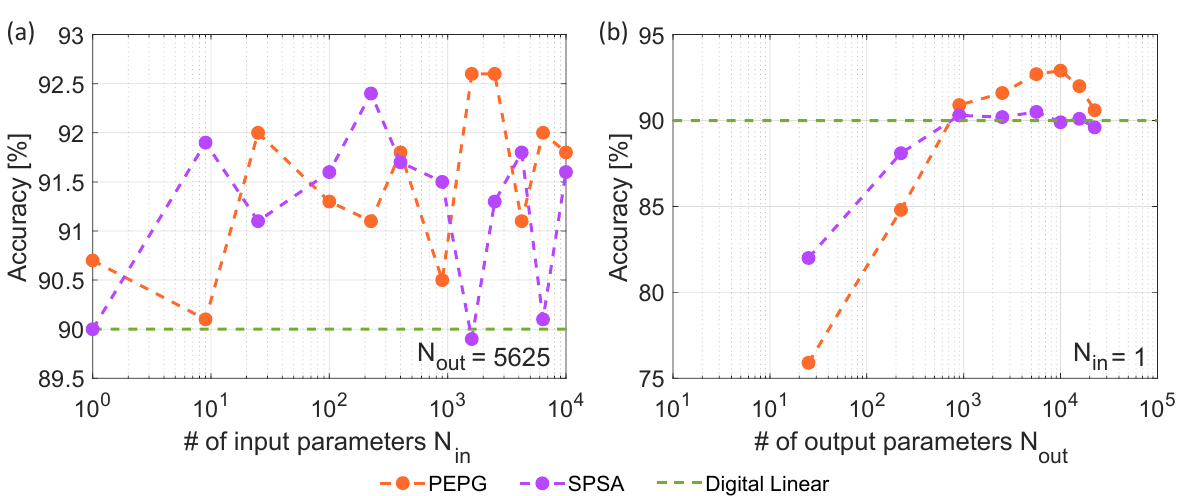}
    \caption{Performance comparison between SPSA (purple) and PEPG (orange) as a function of the number of input (a) and output (b) parameters.}
    \label{fig:in_out_PEPG_SPSA}
    \end{center}
\end{figure}

For a given LA-VCSEL, we can study how performance scales with the number of input and output parameters for, both, SPSA and PEPG as shown in Fig. \ref{fig:in_out_PEPG_SPSA}. Interestingly, for the output weights, both strategies can overcome the digital linear classifier limit, but PEPG is able to consistently reach better performance at high numbers of parameters and it also saturates later at $N_{\text{out}} = 5625$ as opposed to $N_{\text{out}} = 2500$, meaning that it can leverage these new dimensions whereas SPSA cannot. For the input weights, our measurement is very noisy and we would need to conduct a more comprehensive study averaging several times for each datapoint to get more reproducible results, which is currently out of scope due to the slow SLM.

\begin{figure}[h!]
    \begin{center}
    \includegraphics[width=1\linewidth]{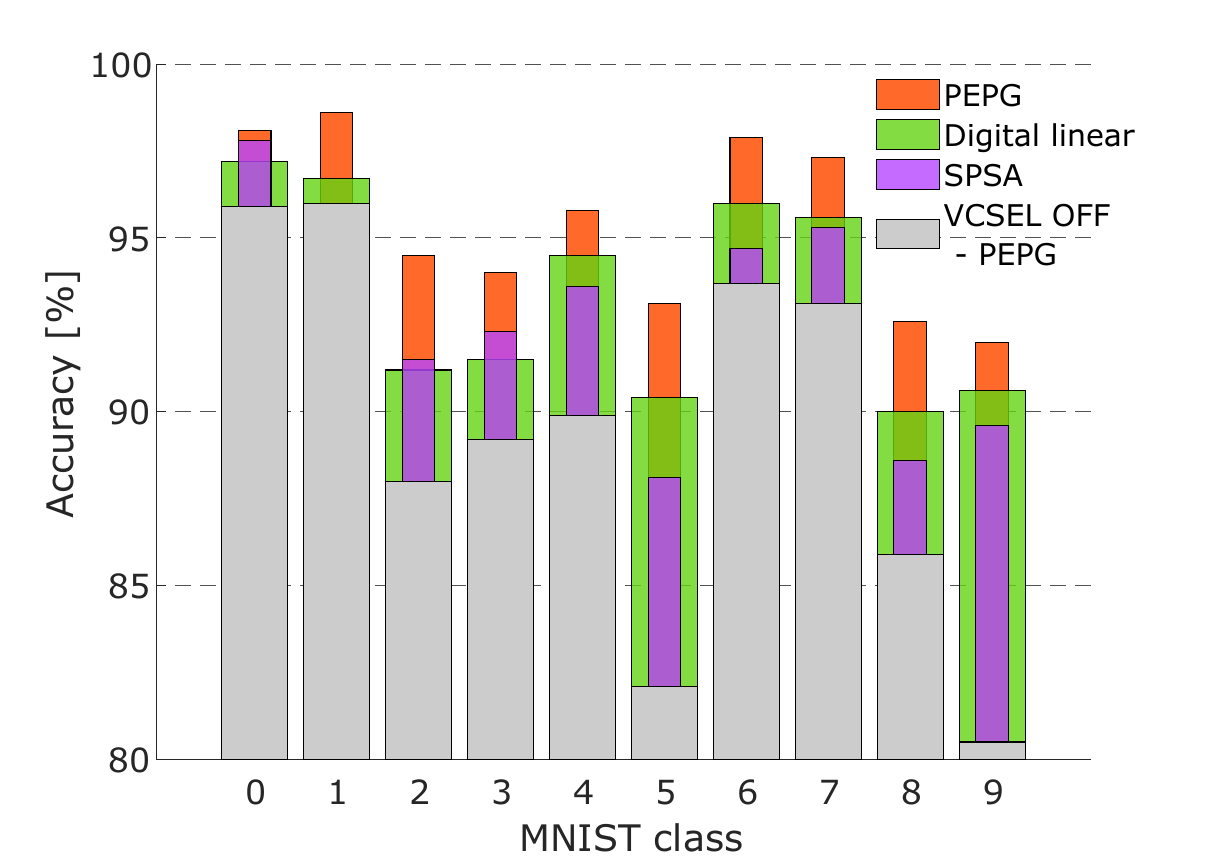}
    \caption{One-vs-all test classification accuracy on the MNIST dataset, using PEPG (orange) and SPSA (purple) compared to a digital linear classifier (green). The VCSEL off baseline is shown for reference (gray).}
    \label{fig:full_classes_VCSEL}
    \end{center}
\end{figure}

Finally, we can benchmark relevant strategies like PEPG and SPSA on all classes of the MNIST dataset in a sequential manner using one-vs-all classification, and compare them to the baseline of a digital linear classifier trained and tested directly on the same Boolean MNIST images that are sent to the LA-VCSEL. In one-vs-all classification, for $n_{\text{classes}}$ classes, $n_{\text{classes}}$ independent classifiers are trained from scratch, each distinguishing one class from the rest.
The results from this comparison are shown in Fig. \ref{fig:full_classes_VCSEL}. 

PEPG marks itself as the most efficient online learning algorithm and converges to a lower error faster than the SPSA algorithm while also providing more consistent behavior. Interestingly, only PEPG is able to overcome the digital linear classifier even if SPSA gets quite close with only a ${\sim} 0.6\%$ difference on average across all classes while PEPG overcomes the digital linear classifier limit by $2\%$.Table \ref{tab:full_classes_VCSEL} shows the averaged classification accuracy for each class of the MNIST dataset for the linear classifier, PEPG and SPSA, as well as with the LA-VCSEL switched off, which constitutes the baseline for a hardware linear system, comprising only the linear random mixing via the MMF. Crucially, when the LA-VCSEL is off, we specifically train the MMF system with PEPG, which is the most efficient and high-performance algorithm. Moreover, we use the same number of input and output parameters for, both, VCSEL ON and OFF. Therefore, the gain in performance when switching the LA-VCSEL on is solely due to its nonlinearity and high dimensional mapping. In principle, even with the LA-VCSEL switched off, there still is a nonlinearity present in the physical system. Indeed, our output detector by measuring the light's intensity $I^{\text{out}} = \sum_{i} W_i^{\text{out}}*|E_i^{\text{MMF}}|^2$, implements a nonlinear operation on the MMF field, yet as shown in our present measurement, this nonlinearity is not sufficient at all and cannot even result in performance close to a digital linear classifier.

\begin{table}[h!]
    \centering
    \begin{tabular}{ |c|c|c|c|c| }
    \hline
    Digital Linear Classifier & PEPG & SPSA & VCSEL OFF PEPG \\
    \hline
    93.37 & 95.39 & 92.71 & 89.43 \\
    \hline
    \end{tabular}
    \caption{Averaged classification accuracy across all classes of the MNIST dataset for the linear classifier, PEPG and SPSA, as well as with the LA-VCSEL switched off.}
    \label{tab:full_classes_VCSEL}
\end{table}

These results might seem trivial or obvious given the LA-VCSEL's highly nonlinear behavior. Yet, because the linear classifier is trained in software on a digital computer, it is completely exempt of noise, instabilities, drifts, and benefits from floating point precision digits. In addition, it is trained in one step with a ridge regression algorithm to prevent overfitting and help its generalization properties. For these reasons, we feel that the fact that a fully hardware trained ONN with no offline components can overcome a digital, albeit linear, classifier is a significant achievement.

\section{Conclusion}

In summary, our work presents a comprehensive exploration of model‐free, hardware‐compatible training algorithms implemented on a laser‐based optical neural network (ONN). The investigation began with a detailed ceiling analysis performed on a representative software neural network, which revealed that the inclusion of both positive and negative weights, at least 4 to 5-bit weight resolution, and tunable input connectivity are fundamental for achieving high performance in complex tasks. In our simulations, a fully trainable feed-forward neural network achieved near 97.5\% accuracy on the MNIST dataset, whereas fixed random mappings as found in extreme learning machines (ELM) resulted in a significantly lower performance 87\%. These observations underscore the importance of trainable input weights and the critical role of weight precision, especially when processing spatially diverse and high-dimensional data.

Building on the insights from this ceiling analysis, we realized significant improvements to our ONN experimental setup. We designed both input and output weights with high resolution as well as positive and negative values, to unlock the full potential of our LA-VCSEL based NN concept. These hardware improvements allowed us to move beyond hardware RC and implement a fully trainable ONN. In addition, we implemented and extensively studied the hyperparameter behavior of different hardware compatible optimization algorithms, namely SPSA, PEPG, CMA-ES and FD, to train our ONN. We also compared the convergence efficiency of these strategies in the context of hardware optimization under limited digital resources for a wide range of LA-VCSELs, and found that PEPG is the most efficient and high-performance algorithm. In contrast, due to its computational overhead, CMA-ES is not well suited for high dimensional optimization problems in an autonomous hardware context. Finally, SPSA offers somewhat of a happy medium between to two. We then studied the performance of PEPG and SPSA for all classes of the MNIST dataset and compared them to two baselines. The first being a digital linear classifier, while the second was our ONN with the LA-VCSEL switched off, which corresponds to a hardware linear system. We found that PEPG is able to overcome the digital linear classifier limit of 93.37\%, while SPSA approaches it yet falls short.
Crucially, both outperform the linear hardware system accuracy of 89.43\% by a significant margin achieving 95.39\% and 92.71\% for PEPG and SPSA respectively, highlighting the high dimensional nature of the transformation produced by the LA-VCSEL. Notably, a significant portion of physical NNs in the literature still rely on offline, i.e. software weights on a digital computer. This approach, while understandable since it offers great flexibility, is not in the spirit of building autonomous computing systems. Crucially, it artificially inflates the performance of physical NNs while tying them down, in most cases, to a slow digital computer handling the states of the system and applying software weights, which undermines the potential gains in terms of speed and energy efficiency that physical systems might otherwise offer. In our view, offline weights should be reserved for fundamental testing and troubleshooting, which is illustrated by the stark performance difference between the digital linear classifier and the VCSEL OFF.

In conclusion, our study not only provides a robust framework for training large-scale physical neural networks but also introduces experimental innovations that overcome longstanding challenges in the field. The combination of model-free optimization, high-resolution tunable weight control, and an advanced VCSEL-based platform culminates in an ONN that is both high-performing and energy-efficient. We hope that these findings will stimulate further advancements in photonic neural computing and foster interdisciplinary collaboration across physics, photonics, and machine learning, ultimately driving the next generation of unconventional computational hardware.

\backmatter

\section*{Acknowledgments}

This work was supported by the Region Bourgogne Franche-Comt\'{e}, the EUR EIPHI program (Contract No. ANR-17-EURE-0002),  by the German Research Foundation (via Phase III of the Collaborative Research Center project 787), the European Union’s and Research Council's Horizon research and innovation program under the Marie Skłodowska-Curie grant agreement No 860830 (POST DIGITAL) and the ERC Consolidator Grant (Grant agreement No. 101044777 (INSPIRE)) as well as JSPS KAKENHI (Grant No. JP22H05198) and JST CREST (Grant No. JPMJCR24R2)

\section*{Data availability}

The data generated and/or analyzed during the current study are not publicly available for legal/ethical reasons but are available from the corresponding author on reasonable request.

\section*{Code availability}
The code used in the simulation section is availble in the following open Github repository : 
\url{https://github.com/ASkalli/learning_strategies}

\section*{Competing interests}
Competing interests: We declare that none of the authors have competing financial or non-financial interests as defined by Nature Portfolio.

\section*{Author contributions}

AS and DB conceived the experiment and in combination with SS and MG implemented the learning algorithms.
MG and TC fabricated and characterized the LA-VCSEL devices devices designed by JL with help from SR.
AS carried out all experiments under the supervision of DB. AS wrote the manuscript with contributions of all authors.


\begin{appendices}




\end{appendices}

\bibliographystyle{ieeetr}

\bibliography{sn-bibliography}

\end{document}